\documentclass{article} 
\usepackage{iclr2027_conference,times}

\usepackage{amsmath,amsfonts,bm}

\def\eqref#1{equation~\ref{#1}}

\def\1{\bm{1}}

\def\vtheta{{\bm{\theta}}}

\DeclareMathAlphabet{\mathsfit}{\encodingdefault}{\sfdefault}{m}{sl}
\SetMathAlphabet{\mathsfit}{bold}{\encodingdefault}{\sfdefault}{bx}{n}

\usepackage{hyperref}
\usepackage{url}
\usepackage{graphicx}
\usepackage{booktabs}
\usepackage{multirow}
\usepackage[normalem]{ulem}
\usepackage{makecell}
\usepackage{wrapfig}
\usepackage{tabularx}
\usepackage{enumitem}
\usepackage[table]{xcolor}
\usepackage{tablefootnote}
\definecolor{originblue}{HTML}{1A1579}
\definecolor{nvidiagreen}{HTML}{76B900}
\definecolor{taured}{HTML}{FF0000}

\DeclareFontEncoding{LS1}{}{}
\DeclareFontSubstitution{LS1}{stix2}{m}{n}
\DeclareSymbolFont{stixsymbols}{LS1}{stix2bb}{m}{it}

\DeclareMathSymbol{\filledheart}{\mathord}{stixsymbols}{"FE}
\DeclareMathSymbol{\filleddiamond}{\mathord}{stixsymbols}{"FF}

\newcommand{\originmark}{%
    \hspace{0.12em}\textsuperscript{\textcolor{originblue}{$\spadesuit$}}%
}
\newcommand{\nvidiamark}{%
    \hspace{0.12em}\textsuperscript{\textcolor{nvidiagreen}{$\filleddiamond$}}%
}
\newcommand{\taumark}{%
    \hspace{0.12em}\textsuperscript{\textcolor{taured}{$\filledheart$}}%
}

\title{%
\mbox{Stuck on Shuffle: On Attribute--Object}\protect\\
\mbox{Binding in Vision--Language Models}%
}

\author{%
\makebox[\dimexpr\textwidth-2\tabcolsep\relax][c]{
    \textbf{Netanel Y. Tamir}\originmark
    \hspace{0.46cm}
    \textbf{Michael Green}\originmark
    \hspace{0.46cm}
    \textbf{Issar Tzachor}\originmark
    \hspace{0.46cm}
    \textbf{Or Shimshi}\originmark
}
\\[3pt]
\makebox[\dimexpr\textwidth-2\tabcolsep\relax][c]{
    \textbf{Dvir Samuel}\nvidiamark
    \hspace{0.46cm}
    \textbf{Or Patashnik}\taumark
    \hspace{0.46cm}
    \textbf{Rami Ben-Ari}\originmark
}
\\[3pt]
\makebox[\dimexpr\textwidth-2\tabcolsep\relax][c]{
    \originmark\,OriginAI
    \hspace{0.46cm}
    \nvidiamark\,NVIDIA
    \hspace{0.46cm}
    \taumark\,Tel Aviv University
}
\\[3pt]
\makebox[\dimexpr\textwidth-2\tabcolsep\relax][c]{
    Project page:
    \href{https://stuck-on-shuffle.github.io/}
         {\texttt{https://stuck-on-shuffle.github.io/}}
}
}

\iclrfinalcopy 
\begin{document}

\maketitle

\vspace{-0.8cm}

\begin{figure*}[h]
    \centering
    \includegraphics[width=\textwidth]{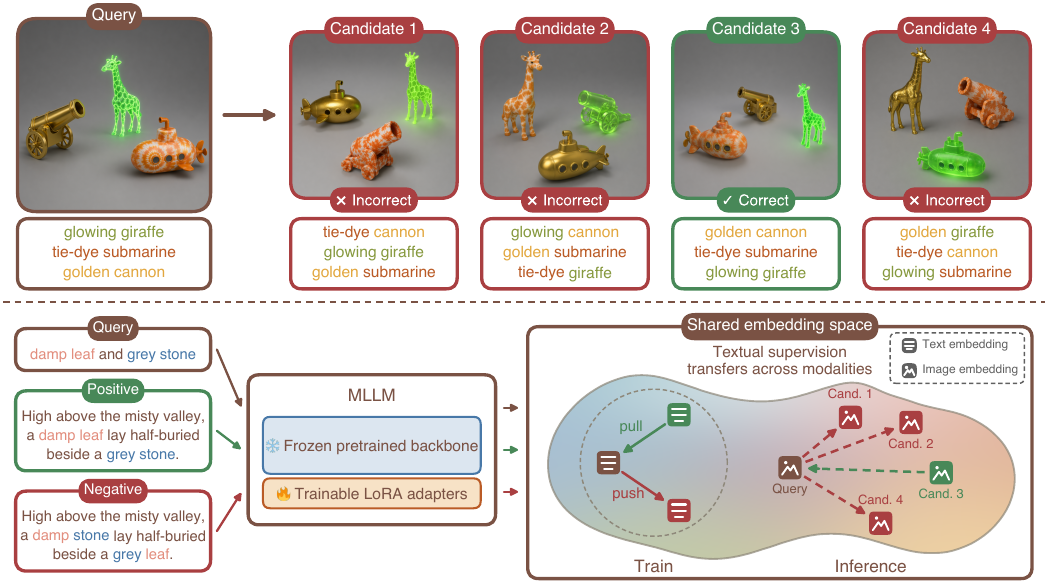}
    \caption{\textbf{Overview.} \textbf{Top:} SABER evaluates attribute--object binding by reassigning fixed sets of objects and attributes. \textbf{Bottom:} Synthetic text-only finetuning improves binding in text and image representations, boosting unimodal and cross-modal retrieval on SABER and existing benchmarks}    
    \label{fig:teaser}
\end{figure*}

\begin{abstract}
\vspace{-0.2cm}

Retrieval is a fundamental task in multimodal systems: given an image or text query, the goal is to find the most relevant images or texts in a collection, often using a shared embedding space. Reliable retrieval requires these embeddings to capture not only which concepts are present, but also how they are composed. One key compositional challenge is \emph{attribute--object binding}: encoding which attributes belong to which objects.
Existing benchmarks provide a limited view of this capability, typically covering a subset of modality pairings, simple two-object swaps, or small attribute sets. Consequently, it remains unclear how robustly current models preserve attribute--object bindings across different modality pairings and more complex settings.
We therefore introduce \textbf{SABER}, a controlled benchmark for evaluating attribute--object binding across all four image and text modality pairings, spanning diverse attribute types and scenes with up to ten objects. Evaluating models on SABER reveals substantial binding failures in both CLIP-style dual encoders and recent MLLM-based embedders.
These findings motivate us to ask how binding can be improved efficiently. While multimodal supervision may be effective, collecting suitable images is difficult and costly. Text, on the other hand, is well suited to binding supervision: it is cheap to generate, supports a broader range of attributes and enables
precisely controlled hard negatives. Recent work also shows that capabilities learned through text-only supervision can transfer to visual modalities in MLLMs. We thus finetune an MLLM-based embedder exclusively on synthetic attribute-binding text. Despite the absence of image supervision, the gains transfer to cross-modal and image--image retrieval while preserving general multimodal embedding performance. Our model achieves SOTA results on SABER 
while also improving on negation and instance-retrieval benchmarks, demonstrating broader benefits of our binding-focused supervision.

\end{abstract}

\section{Introduction}
\label{introduction}

\vspace{-0.2cm}

Multimodal retrieval aims to identify the most relevant images or texts in a large collection for a given image or text query. Modern systems typically use vision--language models (VLMs) to embed different modalities in a shared space~\citep{radford2021learning,jiang2025vlmvec}. Retrieval quality depends on these embeddings encoding both which concepts are present and how those concepts are composed. By encoding collection items once and storing their embeddings, this approach can scale to large collections through efficient similarity search~\citep{johnson2019billion}. In contrast, pairwise models must jointly process the query with each candidate, making them prohibitively expensive at large scale~\citep{reimers2019sentence}.

One important aspect of composition is attribute--object binding: associating each attribute with the correct object~\citep{thrush2022winoground,lewis2024does}. As illustrated in Figure~\ref{fig:teaser} (top), given a query containing a \emph{golden cannon}, \emph{tie-dye submarine}, and \emph{glowing giraffe}, a model should retrieve candidates preserving these bindings rather than those with the same objects and attributes paired differently. Yet VLM embeddings often behave like bag-of-words matchers, capturing which concepts are present without reliably preserving their bindings~\citep{yuksekgonul2022and}.

However, existing evidence provides only a partial picture of these binding failures, since current benchmarks cover a limited range of settings. For example, many benchmarks evaluate only a single modality pairing, such as image queries with text candidates~\citep{yuksekgonul2022and,hsieh2023sugarcrepe}, while others use a limited attribute vocabulary, often focusing on color~\citep{tang2023lemons,burapacheep2024colorswap}, and most focus on simple two-object attribute swaps~\citep{dumpala2024sugarcrepe++,koishigarina2025clip}. This makes it difficult to determine whether these binding failures are general or specific to particular modalities, scene complexity, or attribute types.

To study these questions systematically, we introduce \textbf{SABER} (\textit{\textbf{S}ynthetic \textbf{A}ttribute
\textbf{B}inding \textbf{E}valuation for \textbf{R}epresentations}), a controlled synthetic benchmark that isolates attribute--object binding by holding the set of objects and attributes fixed while varying their assignments. 
\begin{figure*}[h]
    \centering
    \vspace{-0.1cm}
    \includegraphics[width=\textwidth]{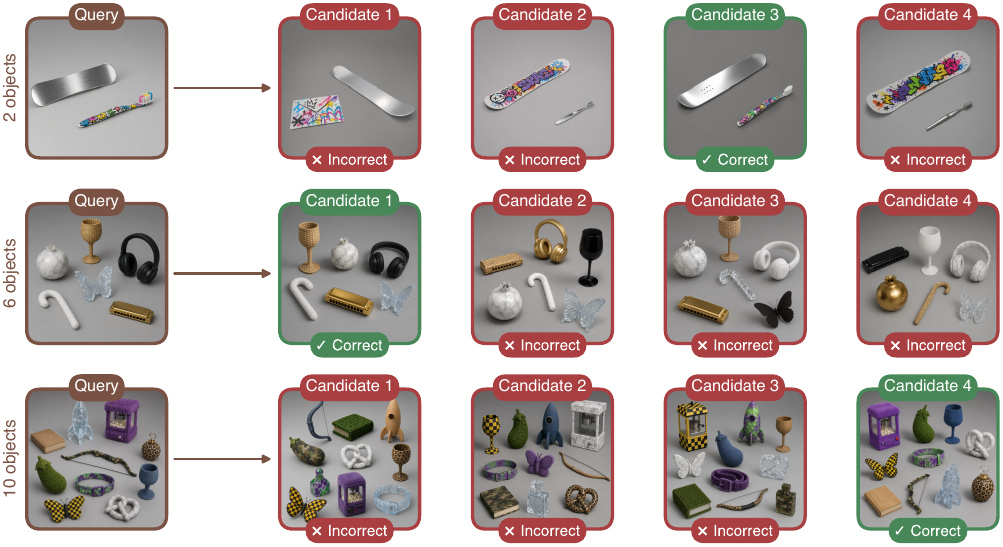}
    \vspace{-0.5cm}
    \caption{\textbf{SABER multiple-choice questions.}
    Each query has four candidates with 2--10 attribute--object pairs: one preserves the bindings, while the others form hard negatives by reassigning attributes. For two-object examples, one negative instead replaces an object or attribute. Although only images are shown, the accompanying captions enable evaluation within and across modalities.}
    \label{fig:dataset_a}
\end{figure*}
SABER supports unimodal and cross-modal retrieval over images and text, with broad object and attribute vocabularies and scenes containing up to ten objects. We use SABER to evaluate dual-encoder VLMs, such as CLIP~\citep{radford2021learning}, as well as newer embedding models derived from multimodal large language models (MLLMs), such as Qwen3-VL-Embedding~\citep{qwen3vlembedding}. Our evaluation on SABER reveals substantial attribute-binding failures across both model families, all four modality pairings, and all levels of scene complexity.

We next ask whether these failures can be reduced through targeted binding supervision. Direct multimodal supervision is a natural option, but constructing controlled images with the same concepts but different attribute--object assignments is difficult and costly. Text offers a simpler alternative: it is cheap to generate, can express visual and non-visual attributes, makes attribute--object bindings explicit rather than implicit in pixels, and enables precisely controlled hard negatives by reassigning attributes among objects. Furthermore, recent studies on MLLMs show that capabilities acquired through text-only supervision can transfer to other modalities~\citep{jiang2024e5,li2025temporal,hu2026praxis}. Together, these observations motivate us to ask whether refining a pretrained MLLM-based multimodal embedder for a specific capability using text alone can transfer improvements across modalities. We test this by finetuning a pretrained multimodal embedder exclusively on synthetic attribute--binding text (Figure~\ref{fig:teaser}, bottom).

Despite receiving no image supervision during finetuning, our model achieves the best overall performance on SABER, substantially outperforming all evaluated baselines across all four modality pairings. The gains extend beyond SABER: our model outperforms existing models on all $9$ attribute-binding benchmarks tested, spanning natural and synthetic images and diverse text sources, while largely preserving general multimodal embedding performance. Beyond attribute binding, finetuning substantially improves performance on both negation and instance retrieval benchmarks, suggesting that binding-focused supervision can yield broader improvements beyond its original training objective. We will release SABER, our textual finetuning dataset, and our finetuned models to support future research on compositionality and attribute binding.

\section{Related Work}
\label{related_work}
\vspace{-0.2cm}

\subsection{Multimodal Embedding Models}
\label{sec:rw-embeddings}


Vision--language embedding models have traditionally relied on dual encoders trained contrastively on web-scale image--caption pairs, as popularized by CLIP~\citep{radford2021learning} and ALIGN~\citep{jia2021scaling}. Subsequent work has advanced this paradigm through greater scale~\citep{cherti2023reproducible}, improved data curation and training recipes~\citep{gadre2023datacomp,sun2023eva}, and alternative objectives such as SigLIP's sigmoid loss~\citep{zhai2023sigmoid}.

More recently, pretrained MLLMs have been adapted into task-conditioned embedding models for multimodal retrieval. VLM2Vec~\citep{jiang2025vlmvec} uses multitask contrastive training to convert a pretrained MLLM into an embedding model. LamRA-Ret~\citep{liu2025lamra} adds language-only retrieval pretraining before finetuning on multimodal retrieval tasks, improving retrieval performance. Qwen3-VL-Embedding~\citep{qwen3vlembedding} combines large-scale contrastive pretraining with distillation from a reranker, which jointly processes each query--candidate pair and assigns it a relevance score. 
The MMEB benchmark series~\citep{jiang2025vlmvec,meng2026vlmvecv,huang2026mmeb} evaluates such multimodal embedders across diverse retrieval settings and input modalities. Our work evaluates attribute--object binding in both CLIP-style dual encoders and MLLM-based embedders.

\subsection{Compositionality in Vision--Language Representations}

Prior work has repeatedly found that VLM embeddings exhibit bag-of-words behavior, with limited sensitivity to compositional structure. Winoground~\citep{thrush2022winoground} illustrates this using captions containing the same words but expressing different compositions, such as \emph{``some plants surrounding a lightbulb''} and \emph{``a lightbulb surrounding some plants,''} which models often struggle to match to the correct images. ARO~\citep{yuksekgonul2022and} and SugarCrepe~\citep{hsieh2023sugarcrepe} probe these failures more systematically using hard negatives that swap or replace objects and attributes, modify relations, or alter word order.

Multiple explanations for this bag-of-words behavior have motivated different mitigation strategies. One line of work attributes
these failures to insufficient compositional signal or other biases in standard
image--text training data
~\citep{doveh2023dense,patel2024tripletclip,gurung2025common}.
Accordingly, proposed remedies include augmenting the training data with compositional hard negatives~\citep{yuksekgonul2022and,zhang2024contrasting,patel2024tripletclip} and improving the caption density and image--text alignment~\citep{doveh2023dense}. However, beyond being costly, automatically constructed multimodal supervision can also be noisy, with artifacts in text perturbations, hallucinations in generated captions, and unintended variation in synthetic images.

A second explanation is that a single global embedding may be too blunt a representation for compositional structure, requiring it to encode diverse information, such as content, appearance, spatial layout, and attribute--object bindings, all within the same vector~\citep{kamath2023text,kang2025clip,weller2025theoretical}. Methods addressing this bottleneck either retain token and patch representations for late-interaction scoring~\citep{kang2025clip,assouel2026object} or use pairwise scorers to directly model interactions between inputs~\citep{krojer2023diffusion,lin2024revisiting,lin2024evaluating}.

We instead use synthetic text-only supervision, where attribute--object bindings can be manipulated precisely and at scale to construct controlled hard negatives, without requiring controlled image generation or changing the standard single-vector architecture and inference procedure.

\subsection{Attribute-Binding Benchmarks}
Existing attribute-binding benchmarks vary considerably in scope and design. VG-Attribution~\citep{yuksekgonul2022and} and SugarCrepe~\citep{hsieh2023sugarcrepe} use natural images but are limited to image--text retrieval. SugarCrepe++~\citep{dumpala2024sugarcrepe++} extends SugarCrepe with text--text retrieval, while ColorSwap~\citep{burapacheep2024colorswap} and BiVLC-Swap-Att.~\citep{miranda2024bivlc} leverage synthetic images to enable bidirectional image--text retrieval.

More controlled benchmarks isolate attribute--object binding through explicitly constructed scenes with limited vocabularies. NCD~\citep{tang2023lemons,anwar2025image} evaluates color binding in scenes containing fruits and vegetables. CLEVR-based benchmarks~\citep{johnson2017clevr,koishigarina2025clip} use rendered geometric objects and scale to scenes with up to ten bindings, while PUG:SPAR~\citep{bordes2023pug} and PUG:SPARE~\citep{koishigarina2025clip} systematically vary attributes in rendered animal scenes. Despite this greater control, evaluation remains limited to image--text retrieval. Figure~\ref{fig:all_datasets} in the Appendix provides representative examples from each benchmark. 

\begin{table}[h]
\centering
\normalsize
\setlength{\tabcolsep}{3.5pt}
\renewcommand{\arraystretch}{1.3}
\setlength{\arrayrulewidth}{0.6pt}
\vspace{-0.4cm}
\caption{
\textbf{Existing attribute-binding benchmarks.}
\textit{Open} denotes an unrestricted vocabulary; MCQ includes Winoground-style $2\times2$ matching and SugarCrepe++'s three-caption evaluation.
}
\label{tab:dataset_comparison}

\resizebox{\linewidth}{!}{%
\begin{tabular}{lcccc|cc|cccc}
\toprule
& \multicolumn{4}{c}{\textbf{Modalities}}
& \multicolumn{2}{c}{\textbf{Evaluation Method}}
& \multicolumn{4}{c}{\textbf{Benchmark Statistics}} \\
\cmidrule(lr){2-5}
\cmidrule(lr){6-7}
\cmidrule(lr){8-11}
\textbf{Dataset}
& \textbf{T$\rightarrow$T}
& \textbf{T$\rightarrow$I}
& \textbf{I$\rightarrow$T}
& \textbf{I$\rightarrow$I}
& \multicolumn{1}{c}{\hspace{6pt}\textbf{MCQ}}
& \multicolumn{1}{c}{\textbf{Retrieval}\hspace{6pt}}
& \textbf{\makecell{\# Eval.\\Samples}}
& \textbf{\makecell{Object\\Vocab.}}
& \textbf{\makecell{Attribute\\Vocab.}}
& \textbf{\makecell{\# Bindings\\ / Sample}} \\
\midrule
NCD
& $\times$
& $\times$
& $\checkmark$
& $\times$
& \hspace{8pt}$\checkmark$
& $\times$
& $1{,}998$
& $19$
& $6$
& $2$ \\

VG-Attribution
& $\times$
& $\times$
& $\checkmark$
& $\times$
& \hspace{8pt}$\checkmark$
& $\times$
& $28{,}748$
& \textit{open}
& \textit{open}
& $2$ \\

ColorSwap
& $\times$
& $\checkmark$
& $\checkmark$
& $\times$
& \hspace{8pt}$\checkmark$
& $\times$
& $1{,}000$
& \textit{open}
& \textit{open}
& $2$ \\

SugarCrepe-Swap-Att.
& $\times$
& $\times$
& $\checkmark$
& $\times$
& \hspace{8pt}$\checkmark$
& $\times$
& $666$
& \textit{open}
& \textit{open}
& $2$ \\

SugarCrepe++-Swap-Att.
& $\checkmark$
& $\times$
& $\checkmark$
& $\times$
& \hspace{8pt}$\checkmark$
& $\times$
& $666$
& \textit{open}
& \textit{open}
& $2$ \\

BiVLC-Swap-Att.
& $\times$
& $\checkmark$
& $\checkmark$
& $\times$
& \hspace{8pt}$\checkmark$
& $\times$
& $278$
& \textit{open}
& \textit{open}
& $2$ \\

CLEVR
& $\times$
& $\times$
& $\checkmark$
& $\times$
& \hspace{8pt}$\checkmark$
& $\times$
& $43{,}519$
& $3$
& $8$
& $2$--$10$ \\

PUG:SPAR
& $\times$
& $\times$
& $\checkmark$
& $\times$
& \hspace{8pt}$\checkmark$
& $\times$
& $19{,}840$
& $32$
& $4$
& $2$ \\

PUG:SPARE
& $\times$
& $\times$
& $\checkmark$
& $\times$
& \hspace{8pt}$\checkmark$
& $\times$
& $88{,}704$
& $12$
& $8$
& $2$ \\

\midrule

SABER (ours)
& $\checkmark$
& $\checkmark$
& $\checkmark$
& $\checkmark$
& \hspace{8pt}$\checkmark$
& $\checkmark$
& $1{,}800$
& $232$
& $50$
& $2$--$10$ \\

\bottomrule
\end{tabular}%
}

\end{table}

\section{SABER Benchmark}
\label{sec:dataset}
\vspace{-0.2cm}

Existing attribute-binding benchmarks are limited in modality coverage, vocabulary, or the number of bindings per sample, and rely on small candidate-set evaluations without assessing large-gallery retrieval with multiple valid positives~\citep{yuksekgonul2022and,hsieh2023sugarcrepe,burapacheep2024colorswap,miranda2024bivlc,koishigarina2025clip}. As a result, they provide only a partial view of how well current representations preserve attribute--object bindings.

To address these limitations, we introduce \textbf{SABER} (\textit{\textbf{S}ynthetic \textbf{A}ttribute \textbf{B}inding \textbf{E}valuation for \textbf{R}epresentations}), a benchmark structured as 4-way multiple-choice questions (MCQs), whose candidates can also be pooled for large-gallery retrieval. SABER covers all four image and text modality pairings, draws from sets of $232$ everyday objects and $50$ visual attributes spanning colors, materials, patterns, and effects, and scales to scenes with up to ten objects. Table~\ref{tab:dataset_comparison} summarizes the differences between SABER and prior benchmarks.

Each of our MCQs, illustrated in Figures~\ref{fig:teaser} and~\ref{fig:dataset_a}, contains a query and four candidates, each represented by both an image and a caption. One candidate preserves the query's attribute--object bindings, while the remaining three are hard negatives that alter these assignments. Except for one negative in the two-object setting (Candidate~1 in Figure~\ref{fig:dataset_a}), all candidates contain the same objects and attributes as the query, isolating binding from other challenges such as object recognition.

The benchmark draws from our sets of $232$ everyday objects and $50$ visual attributes. Sampled objects and attributes are paired in a templated query caption, such as \textit{``a sandy snowboard and a pink cherry.''} This fixed phrasing minimizes linguistic variation unrelated to attribute--object binding. After constructing the corresponding candidate captions, we synthesize the query and candidate images using GPT~Image~2~\citep{openai2026chatgptimages2}, with instructions specifying neutral CLEVR-style scenes and consistent realizations of each attribute across all five images in an MCQ. This keeps attribute appearance consistent across candidates, so that their visual differences primarily reflect changes in binding. All generated images are manually inspected to ensure that the intended objects and attribute bindings are rendered correctly, yielding unambiguous MCQs with a unique correct answer.

\begin{figure*}[t]
    \centering
    \includegraphics[width=\textwidth]{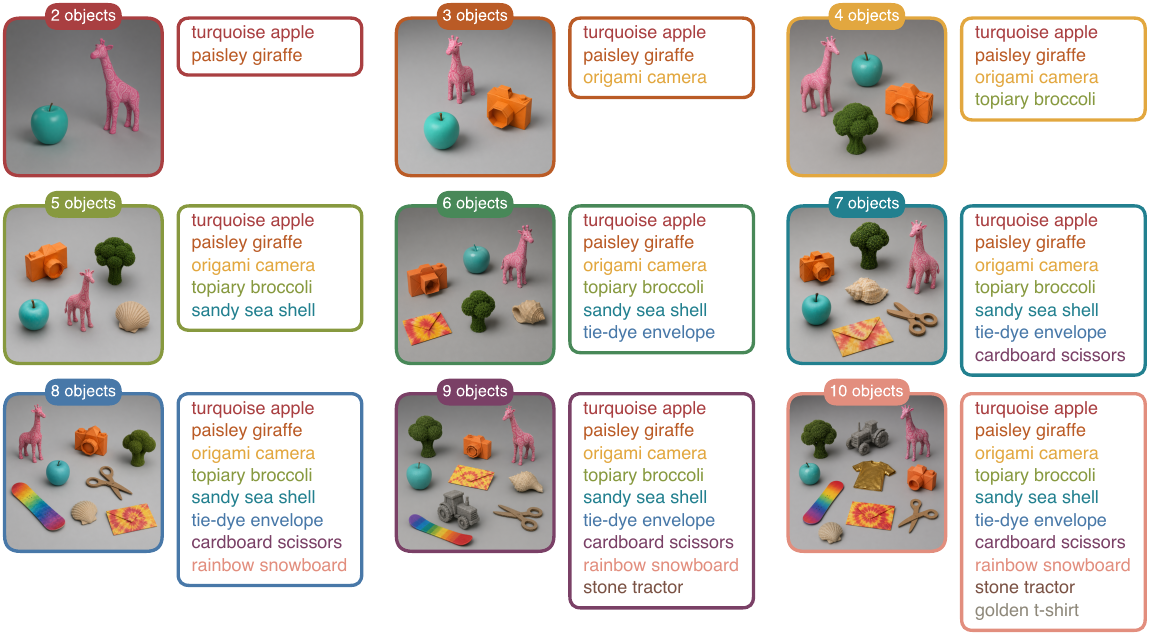}
    \vspace{-0.6cm}
    \caption{\textbf{Nested SABER chains.}
    Nine linked MCQs form a chain from two to ten objects, read left to right and top to bottom. Each successive query adds one binding while preserving all earlier ones. Half of SABER's MCQs are organized into such chains, allowing later correct candidates to serve as additional positives for earlier queries in large-gallery retrieval.}
    \label{fig:dataset_b}
    \vspace{-0.4cm}
\end{figure*}

SABER contains $200$ MCQs for each scene size from $2$ to $10$ objects: $100$ independent examples and $100$ arranged into nested chains, as illustrated in Figure~\ref{fig:dataset_b}. Each successive query in a chain adds one attribute--object binding while preserving earlier ones. Overall, this yields $1{,}800$ queries and $7{,}200$ candidates, or $9{,}000$ image--caption pairs.

We evaluate models on SABER using two complementary protocols: 4-way MCQ and large-gallery retrieval. In the MCQ setting, the model ranks four candidates by similarity to the query and succeeds when the binding-preserving candidate ranks first. In the retrieval setting, each query is ranked against all $7{,}200$ candidates, with any candidate containing all query bindings treated as positive. Independent queries have one positive, while nested-chain queries may have multiple positives from later scenes that preserve earlier bindings. This reflects practical retrieval, where the queried bindings may appear within a larger, more complex scene. Further benchmark-generation details appear in Appendix~\ref{app:dataset-details}.

\section{Text-Only Finetuning for Attribute Binding}
\label{sec:finetuning}
\vspace{-0.2cm}

To assess the binding capabilities of current embedding models, we evaluate a range of CLIP-style dual encoders~\citep{radford2021learning,bolya2025PerceptionEncoder}, including compositional and attribute-binding variants such as TripletCLIP~\citep{patel2024tripletclip}, READ-CLIP~\citep{kwon2026enhancing}, and LABCLIP~\citep{koishigarina2025clip} together with recent MLLM-based embedders~\citep{liu2025lamra,qwen3vlembedding}. Table~\ref{tab:saber_main} reveals clear binding failures across existing models on SABER, motivating targeted finetuning to improve binding sensitivity.

We therefore turn to text-only finetuning on synthetic attribute-binding data. Text is particularly well suited to this setting because bindings are explicit and easy to manipulate, enabling precisely controlled hard negatives without the incidental variation that can arise in generated images.


Using Gemma 4~\citep{gemmateam2026gemma4}, we construct $200{,}000$ finetuning triplets, each containing a query, a positive, and a hard negative. Each query specifies two or three attribute--object pairs, with the positive then contextualizing those bindings. For example, the query \emph{``red banner and torn flag''} may be paired with \emph{``High above the roar of the crowd, \ldots there waved a red banner and a torn flag.''} We construct a hard negative by permuting either the attributes or the objects, so that it differs from the positive only in its attribute--object assignments. To reduce shortcut cues, we choose attributes that remain plausible for all objects in each example and randomize the order of the attribute--object pairs between the query and the positive.

Although the finetuning triplets are entirely textual, our goal is to improve attribute binding across all text and image modality pairings. More formally, let $x_q^m$ denote a query expressing a set of attribute--object bindings in modality $m \in \{\mathrm{Text},\mathrm{Image}\}$, and
let $x_+^n$ and $x_-^n$ denote a binding-preserving positive and a
binding-conflicting negative, respectively, in modality $n \in \{\mathrm{Text},\mathrm{Image}\}$.
Ideally, the embedding model should place the positive closer to the query under every modality
pairing:
\[
D\!\left(f_{\vtheta}(x_q^m), f_{\vtheta}(x_+^n)\right)
>
D\!\left(f_{\vtheta}(x_q^m), f_{\vtheta}(x_-^n)\right),
\qquad
\forall (m,n) \in \{\mathrm{Text},\mathrm{Image}\}^2,
\]
where $f_{\vtheta}$ is the multimodal embedder and $D$ is cosine similarity in the shared embedding space.

We use text-only supervision to encourage this behavior. Given a textual query $t_q$, a positive $t_+$, and a hard
negative $t_-$, we minimize the following triplet loss~\citep{schroff2015facenet}
\[
\mathcal{L}_{\mathrm{text}}
=
\left[
D\!\left(f_{\vtheta}(t_q), f_{\vtheta}(t_-)\right)
-
D\!\left(f_{\vtheta}(t_q), f_{\vtheta}(t_+)\right)
+
\gamma
\right]_+,
\]
where $\gamma$ is the triplet margin and $[z]_+ = \max(z,0)$ denotes the positive-part function.

Unlike CLIP-style dual encoders with separate image and text encoders, MLLMs contain parameters that are shared across modalities. Although we optimize only the textual loss above, updating these shared parameters from textual triplets also affects the representations produced for images, providing a mechanism for the learned binding behavior to transfer to the other modality. We finetune Qwen3-VL-Embedding-2B and Qwen3-VL-Embedding-8B~\citep{qwen3vlembedding} with LoRA~\citep{hu2022lora} using this objective and subsequently apply LiNeS~\citep{wang2025lines} to rescale the layer-wise parameter updates and mitigate forgetting of pretrained capabilities.

Appendix~\ref{app:finetuning-dataset} provides additional details on the finetuning dataset and optimization procedure. Appendix~\ref{app:ablations} ablates the finetuning data and LiNeS and analyzes attention changes before and after finetuning.  Appendix~\ref{app:diff_backbones} shows that our finetuning recipe extends to other MLLM backbones. Appendix~\ref{app:text_vs_images} compares finetuning with text versus images on SABER and shows that textual supervision yields stronger cross-modal retrieval performance.

\section{Experiments}
\label{results}
\vspace{-0.2cm}
\subsection{Can Text-Only Finetuning Improve Attribute Binding Across Modalities?}



Table~\ref{tab:saber_main} compares our finetuned models with several baselines, on SABER under both 4-way multiple-choice questions and large-gallery retrieval. For all models, candidates are ranked by cosine similarity between the corresponding query and candidate embeddings. We report results for all four image and text modality pairings, with retrieval evaluated separately on the independent and nested queries. Among the pretrained models, Qwen3-VL-Embedding-8B achieves the strongest mean performance.

\definecolor{rowgray}{HTML}{EFEEEC}

\begin{table}[t]
\centering
\small
\caption{
\textbf{Performance on SABER.}
SABER-MCQ reports mean 4-way accuracy (\%) and SABER-Retrieval mean mAP (\%), both averaged over 2--10 objects. Retrieval is split into independent and nested subsets over the pooled candidate gallery.
Our finetuned 8B model performs best across all four modality pairings in both protocols.
Results by number of objects are reported in Tables~\ref{tab:saber_mcq_accuracy_full}, \ref{tab:saber_retrieval_map_independent}, and \ref{tab:saber_retrieval_map_nested} in the Appendix.
Best and second-best results are shown in bold and underlined, respectively.
}
\label{tab:saber_main}
\setlength{\tabcolsep}{3pt}
\renewcommand{\arraystretch}{1.15}
\setlength{\arrayrulewidth}{0.6pt}

\resizebox{\linewidth}{!}{%
\begin{tabular}{@{}lccccc|ccccc|ccccc@{}}
\toprule
&
\multicolumn{5}{c|}{\textbf{SABER-MCQ}}
& \multicolumn{5}{c|}{\textbf{SABER-Retrieval (Independent)}}
& \multicolumn{5}{c}{\textbf{SABER-Retrieval (Nested)}} \\
\cmidrule(lr){2-6}
\cmidrule(lr){7-11}
\cmidrule(lr){12-16}

\textbf{Model}
& \textbf{T$\to$T}
& \textbf{T$\to$I}
& \textbf{I$\to$T}
& \textbf{I$\to$I}
& \textbf{Mean}
& \textbf{T$\to$T}
& \textbf{T$\to$I}
& \textbf{I$\to$T}
& \textbf{I$\to$I}
& \textbf{Mean}
& \textbf{T$\to$T}
& \textbf{T$\to$I}
& \textbf{I$\to$T}
& \textbf{I$\to$I}
& \textbf{Mean} \\
\midrule

\rowcolor{rowgray}
CLIP ViT-B/32
& 53.6 & 23.0 & 22.3 & 75.5 & 43.6
& 74.0 & 38.8 & 33.1 & 85.1 & 57.8
& 38.0 & 13.7 & 16.1 & 49.3 & 29.3 \\

LABCLIP ViT-B/32
& 55.2 & 22.8 & 24.2 & 75.3 & 44.4
& 74.3 & 35.0 & 25.8 & 85.1 & 55.0
& 38.1 & 13.0 & 12.1 & 49.2 & 28.1 \\

\rowcolor{rowgray}
TripletCLIP ViT-B/32
& 42.9 & 27.1 & 22.2 & 60.2 & 38.1
& 65.9 & 8.8 & 5.0 & 63.2 & 35.7
& 35.8 & 5.4 & 3.5 & 32.1 & 19.2 \\

READ-CLIP ViT-B/32
& 47.3 & 23.9 & 23.7 & 82.3 & 44.3
& 68.4 & 33.5 & 23.6 & 89.7 & 53.8
& 37.7 & 10.7 & 13.1 & \underline{54.1} & 28.9 \\

\rowcolor{rowgray}
PE-Core-bigG-14-448
& 66.1 & 29.7 & 27.6 & 67.6 & 47.7
& 81.1 & 56.0 & 55.5 & 81.5 & 68.5
& 41.5 & 23.6 & \underline{32.9} & 47.1 & 36.3 \\

LamRA-Ret
& 49.1 & 30.6 & 27.0 & 64.9 & 42.9
& 70.4 & 56.8 & 51.5 & 79.2 & 64.5
& 37.3 & 27.3 & 28.0 & 43.7 & 34.1 \\

\rowcolor{rowgray}
Qwen3-VL-Embedding-2B
& 35.0 & 27.8 & 26.4 & 73.9 & 40.8
& 60.1 & 54.0 & 55.0 & 85.5 & 63.6
& 34.0 & 28.1 & 30.9 & 47.3 & 35.1 \\

Qwen3-VL-Embedding-8B
& 53.7 & \underline{34.9} & \underline{36.3} & \underline{84.9} & 52.5
& 73.5 & \underline{59.6} & \underline{62.6} & \underline{91.9} & 71.9
& 37.8 & \underline{33.1} & 29.9 & 52.3 & 38.3 \\

\midrule

\rowcolor{rowgray}
\textbf{Ours-2B}
& \underline{76.9} & 34.8 & 32.9 & 80.9 & \underline{56.4}
& \underline{87.8} & 59.3 & 58.3 & 89.8 & \underline{73.8}
& \underline{46.2} & 29.4 & 32.5 & 49.9 & \underline{39.5} \\

\textbf{Ours-8B}
& \textbf{99.3} & \textbf{69.0} & \textbf{74.1} & \textbf{94.3} & \textbf{84.2}
& \textbf{99.6} & \textbf{82.3} & \textbf{85.5} & \textbf{97.1} & \textbf{91.1}
& \textbf{65.5} & \textbf{46.3} & \textbf{46.0} & \textbf{58.6} & \textbf{54.1} \\

\bottomrule
\end{tabular}%
}
\vspace{-0.4cm}
\end{table}

Our textual finetuning substantially improves MCQ performance on SABER across all modality pairings. Our finetuned 8B model reaches a mean accuracy of $84.2\%$, up from $52.5\%$ for Qwen3-VL-Embedding-8B, with near-perfect text--text accuracy of $99.3\%$. 
Despite using no visual data during finetuning, it gains $34.1$, $37.8$, and $9.4$ points in the text--image, image--text, and image--image settings, respectively. Our 2B model shows the same pattern, improving over its pretrained counterpart across all four pairings, with an average gain of $15.6$ points.

In independent retrieval, models must find the binding-preserving match within the full 7,200-candidate gallery, containing both binding-swapped distractors and thousands of unrelated scenes. For all models except the CLIP-based ViT-B/32 variants, the correct candidate and its three binding-swapped distractors occupy the top four retrieved positions in over $90\%$ of queries, so retrieval is largely governed by the same binding comparison as the MCQ task. For the CLIP-based models, this occurs in only $27$--$53\%$ of queries overall, with the failure concentrated in the cross-modal pairings. Unrelated scenes are therefore frequently retrieved instead, revealing an additional failure mode not captured by MCQ.  Our 8B model again performs best, improving mean mAP by $19.2$ points over Qwen3-VL-Embedding-8B. 

Retrieval performance drops substantially on the nested subset for all models, reflecting the more demanding setting in which queries may match larger scenes containing additional bindings. Our 8B model nevertheless performs best across all four modality pairings, reaching $54.1\%$ mean mAP compared with $38.3\%$ for Qwen3-VL-Embedding-8B. The nested setting exposes a challenge absent from MCQ and independent retrieval: \emph{containment}, or recognizing valid supersets rather than only exact matches. Across models, candidates containing only a subset of the query bindings rank above the exact match in only $3.3$--$17.1\%$ of comparisons on average across modalities, but above candidates that contain all query bindings plus additional ones in $33$--$39\%$. These results highlight containment as a remaining challenge distinct from binding-swap discrimination.

Tables~\ref{tab:saber_mcq_accuracy_full}, \ref{tab:saber_retrieval_map_independent}, and \ref{tab:saber_retrieval_map_nested} in the Appendix provide results broken down by the number of attribute--object bindings. Notably, the improvements extend to scenes with up to ten objects, even though our finetuning triplets permute bindings among only two or three attribute--object pairs.
\subsection{Do the Gains Generalize Beyond SABER?}

We next test whether the improvements observed on SABER transfer to existing
attribute-binding benchmarks.
Table~\ref{tab:existing_binding_benchmarks} evaluates $9$ existing benchmarks across $12$ retrieval settings spanning natural and synthetic images as well as human-written, LLM-generated, and templated text. We report accuracy under each benchmark's native evaluation protocol. Our finetuned 8B model outperforms every baseline across all $12$ settings, improving over Qwen3-VL-Embedding-8B by $15.4$ points on average. Relative to Qwen3-VL-Embedding-8B, gains reach $22.8$ points in the
SugarCrepe++ text--text setting, $33.9$ points in the ColorSwap text--image setting, and $25.6$ points in the CLEVR
image--text setting. Our finetuned 2B model likewise improves over its pretrained counterpart by $8.2$ points on average.
These results indicate that the learned improvements in attribute binding are not limited to SABER's synthetic scenes or templated text, nor to the structure of the finetuning data, where short queries are matched against contextualized positives and hard negatives.

\vspace{-0.2cm}
\definecolor{rowgray}{HTML}{EFEEEC}

\newcolumntype{C}{>{\fontsize{10.95}{11.5}\selectfont\centering\arraybackslash}m{1.45cm}}
\newcolumntype{Z}{>{\fontsize{10.95}{11.5}\selectfont\centering\arraybackslash}m{1.70cm}}

\newcolumntype{A}{>{\fontsize{10.95}{11.5}\selectfont\hspace*{-10pt}\centering\arraybackslash}m{1.45cm}}
\newcolumntype{B}{>{\fontsize{10.95}{11.5}\selectfont\hspace*{-5pt}\centering\arraybackslash}m{1.70cm}}
\newcolumntype{D}{>{\fontsize{10.95}{11.5}\selectfont\hspace*{-4pt}\centering\arraybackslash}m{1.45cm}}

\newcommand{\modelhead}[2]{%
  {\footnotesize\shortstack[c]{\strut #1\\[-1.5pt]\strut #2}}%
}

\newcommand{\modelheadsingle}[1]{%
  {\footnotesize #1}%
}

\newcommand{\modelheadours}[1]{%
  {\footnotesize\bfseries #1}%
}

\begin{table}[h]
\centering
\caption{
\textbf{Performance on existing attribute-binding benchmarks.}
Accuracy (\%) under each benchmark's native multiple-choice protocol.
Best and second-best results are shown in bold and underlined, respectively.
}
\label{tab:existing_binding_benchmarks}

\normalsize
\setlength{\tabcolsep}{1.0pt}
\renewcommand{\arraystretch}{1.10}

\resizebox{\linewidth}{!}{%
\begin{tabular}{@{}>{\fontsize{10.95}{11.5}\selectfont}lCABZCDDCCC@{}}

\toprule
\textbf{Dataset}
& \modelhead{CLIP}{B/32}
& \modelhead{LABCLIP}{B/32}
& \modelhead{TripletCLIP}{B/32}
& \modelhead{READ-CLIP}{B/32}
& \modelhead{PE}{CORE-G}
& \modelhead{LamRA-}{Ret}
& \modelhead{Qwen3-VL}{Emb-2B}
& \modelhead{Qwen3-VL}{Emb-8B}
& \modelheadours{Ours-2B}
& \modelheadours{Ours-8B} \\

\midrule

\multicolumn{11}{l}{\textbf{Text$\boldsymbol{\rightarrow}$Text}} \\
\addlinespace[2pt]

SugarCrepe++-Swap-Att.
& $32.7$
& $41.0$
& $34.7$
& $56.8$
& $53.2$
& $66.2$
& $50.3$
& $70.9$
& $\underline{78.7}$
& $\mathbf{93.7}$ \\

\midrule

\multicolumn{11}{l}{\textbf{Text$\boldsymbol{\rightarrow}$Image}} \\
\addlinespace[2pt]

\rowcolor{rowgray}
ColorSwap
& $17.9$
& $15.5$
& $14.0$
& $24.5$
& $44.6$
& $23.1$
& $22.8$
& $\underline{45.9}$
& $34.2$
& $\mathbf{79.8}$ \\

BiVLC-Swap-Att.
& $19.1$
& $24.1$
& $7.2$
& $\underline{31.7}$
& $19.4$
& $21.2$
& $15.1$
& $21.6$
& $16.2$
& $\mathbf{32.0}$ \\

\midrule

\multicolumn{11}{l}{\textbf{Image$\boldsymbol{\rightarrow}$Text}} \\
\addlinespace[2pt]

\rowcolor{rowgray}
NCD
& $71.1$
& $66.2$
& $62.7$
& $63.2$
& $70.2$
& $48.3$
& $52.5$
& $\underline{80.6}$
& $60.8$
& $\mathbf{95.4}$ \\

VG-Attribution
& $62.5$
& $69.3$
& $57.6$
& $\underline{80.1}$
& $67.9$
& $76.4$
& $67.0$
& $78.0$
& $67.0$
& $\mathbf{80.4}$ \\

\rowcolor{rowgray}
ColorSwap
& $38.2$
& $29.3$
& $39.6$
& $34.2$
& $67.1$
& $46.4$
& $47.6$
& $\underline{71.4}$
& $56.7$
& $\mathbf{88.4}$ \\

SugarCrepe-Swap-Att.
& $64.0$
& $74.8$
& $64.3$
& $82.7$
& $78.2$
& $79.7$
& $77.2$
& $\underline{85.7}$
& $80.3$
& $\mathbf{90.8}$ \\

\rowcolor{rowgray}
SugarCrepe++-Swap-Att.
& $44.9$
& $53.9$
& $42.0$
& $66.2$
& $59.5$
& $62.5$
& $55.1$
& $65.8$
& $\underline{67.9}$
& $\mathbf{82.7}$ \\

BiVLC-Swap-Att.
& $47.1$
& $43.9$
& $38.1$
& $43.9$
& $69.1$
& $56.8$
& $55.0$
& $\underline{73.7}$
& $59.0$
& $\mathbf{74.8}$ \\

\rowcolor{rowgray}
CLEVR
& $50.3$
& $50.6$
& $51.0$
& $51.1$
& $51.4$
& $52.5$
& $52.1$
& $60.6$
& $\underline{64.1}$
& $\mathbf{86.2}$ \\

PUG:SPAR
& $47.5$
& $52.5$
& $65.7$
& $43.0$
& $46.8$
& $40.9$
& $61.1$
& $48.9$
& $\mathbf{70.3}$
& $\underline{67.7}$ \\

\rowcolor{rowgray}
PUG:SPARE
& $50.4$
& $50.4$
& $50.5$
& $51.5$
& $50.8$
& $51.6$
& $\underline{53.7}$
& $52.8$
& $52.3$
& $\mathbf{69.1}$ \\

\bottomrule
\end{tabular}%
}
\end{table}

\subsection{How Much General Multimodal Performance Is Lost After Finetuning?}

MLLM-based embedders combine modality-specific input components with a largely shared backbone, mapping inputs from multiple modalities into a common embedding space. This enables a single model to support a broad range of retrieval tasks.
A natural concern is that binding-focused finetuning may compromise this broader
multimodal capability. We evaluate this trade-off on the non-audio portion of
MMEB-v3~\citep{huang2026mmeb}, spanning $182$ image, video, visual-document,
text, and agent tasks. As shown in Table~\ref{tab:mmeb_v3}, despite using no MMEB-v3 data during finetuning, aggregate performance changes by only $-0.5$ points for the 2B model and $+0.5$ points for the 8B model. Performance also remains similar across domains, with modest declines on image, video,
and text tasks, mixed changes on visual-document tasks, and gains on agent tasks.

\vspace{-0.3cm}
\begin{table}[h]
\centering
\small
\caption{
\textbf{MMEB-v3 results.}
Average Hit@1 is reported for image, video, and agent tasks and NDCG@5 for VisDoc and text tasks.
Our finetuned models remain within $0.5$ points of their pretrained counterparts overall.
Best results per pretrained--finetuned pair are shown in bold.
}
\label{tab:mmeb_v3}

\scalebox{0.9}{%
\begin{tabular}{lcccccc}
\toprule
\textbf{Model}
& \textbf{Image}
& \textbf{Video}
& \textbf{VisDoc}
& \textbf{Text}
& \textbf{Agent}
& \textbf{All} \\
\midrule

\# of Datasets
& $37$
& $18$
& $27$\tablefootnote{MMEB-v3 reports $24$ VisDoc tasks but enumerates $27$; we score all $27$ for every model.}
& $53$
& $47$
& $182$ \\

\midrule

Qwen3-VL-Embedding-2B
& $\mathbf{69.7}$
& $\mathbf{56.3}$
& $\mathbf{74.2}$
& $\mathbf{40.3}$
& $39.8$
& $\mathbf{52.8}$ \\

\textbf{Ours-2B}
& $67.7$
& $56.1$
& $73.0$
& $39.7$
& $\mathbf{41.1}$
& $52.3$ \\

\midrule

Qwen3-VL-Embedding-8B
& $\mathbf{73.7}$
& $\mathbf{59.6}$
& $74.3$
& $\mathbf{43.7}$
& $39.1$
& $54.7$ \\

\textbf{Ours-8B}
& $72.5$
& $58.7$
& $\mathbf{75.4}$
& $43.1$
& $\mathbf{42.3}$
& $\mathbf{55.2}$ \\

\bottomrule
\end{tabular}%
}
\end{table}
\vspace{-0.2cm}

\section{From Attribute Binding to Negation}
\label{sec:negation}
\vspace{-0.2cm}

Attribute binding is only one form of compositional sensitivity. To assess whether binding-focused finetuning improves compositional sensitivity more broadly, we evaluate our models on NegBench~\citep{alhamoud2025vision}, which tests negation understanding. We compare the base Qwen3-VL-Embedding-8B, our finetuned model, and a variant trained after removing the $7.1\%$ of finetuning
examples containing negation terms such as \emph{no}, \emph{without}, and \emph{neither}. This ablation tests whether gains arise from incidental exposure to negation rather than transfer from binding supervision. We also include Qwen3-VL-Reranker-8B, a stronger pairwise model, as an additional reference.

NegBench groups MCQs according to the type of the correct statement: \emph{affirmation} describes present objects, \emph{negation}
absent objects, and \emph{hybrid} both. Each MCQ contains the correct answer together with one additional candidate of each type. Table~\ref{tab:negbench} shows that finetuning improves overall accuracy by $9.3$ points on the
COCO~\citep{lin2014microsoft} subset and $10.5$ points on the PASCAL VOC 2007~\citep{everingham2010pascal} subset, driven mainly
by large gains on hybrid cases. In these cases, the model must distinguish candidates such as \emph{``bowl but no cup''} and \emph{``cup but no bowl,''} which share the same bag of words but assign negation to different
objects. This closely resembles our attribute-binding setup, where correctness depends on associating the correct property with each object. By contrast, in the affirmation and negation groups, the two same-type
candidates contain different bags of words. Notably, the negation-filtered variant retains nearly all of these gains, suggesting transfer from binding supervision rather than direct negation exposure. Qwen3-VL-Reranker-8B remains stronger on pure-negation cases, but is comparable overall to our model.

\vspace{-0.2cm}
\begin{table}[h]
\centering
\small
\caption{\textbf{NegBench results.} MCQ accuracy (\%) by correct-statement type on the COCO and VOC 2007 subsets. Our finetuned model improves overall performance, with the largest gains on hybrid statements. Best results are shown in bold.
}\label{tab:negbench}
\setlength{\tabcolsep}{4pt}
\renewcommand{\arraystretch}{1.15}

\resizebox{0.9\linewidth}{!}{%
\begin{tabular}{@{}l*{8}{c}@{}}
\toprule
&
\multicolumn{4}{c}{\textbf{NegBench$_{coco}$}}
& \multicolumn{4}{c}{\textbf{NegBench$_{voc}$}} \\
\cmidrule(lr){2-5}
\cmidrule(lr){6-9}
\textbf{Model}
& \textbf{Affirmation}
& \textbf{Negation}
& \textbf{Hybrid}
& \textbf{All}
& \textbf{Affirmation}
& \textbf{Negation}
& \textbf{Hybrid}
& \textbf{All} \\
\midrule
Qwen3-VL-Embedding-8B
& $78.3$ & $6.8$ & $36.3$ & $41.4$
& $89.9$ & $4.7$ & $58.0$ & $39.8$ \\

Ours-8B (negation-filtered)
& $82.7$ & $6.6$ & $58.1$ & $50.3$
& $91.5$ & $6.4$ & $78.2$ & $49.7$ \\

\textbf{Ours-8B}
& $\mathbf{83.4}$ & $6.8$ & $\mathbf{58.5}$ & $\mathbf{50.7}$
& $91.8$ & $6.1$ & $\mathbf{79.6}$ & $50.3$ \\
\midrule
Qwen3-VL-Reranker-8B
& $80.6$ & $\mathbf{17.8}$ & $46.9$ & $49.3$
& $\mathbf{95.0}$ & $\mathbf{28.1}$ & $58.0$ & $\mathbf{50.4}$ \\
\bottomrule
\end{tabular}%
\vspace{-0.1cm}
}
\end{table}
\vspace{-0.2cm}

\section{Broader Benefits: Instance-Level Retrieval}
\label{sec:instance_ret}
\vspace{-0.2cm}


We next ask whether binding-focused finetuning benefits instance retrieval, where the goal is to retrieve the same object instance rather than its category. Instances are often distinguished by their attributes, such as the color of a shirt, suggesting that improved binding could strengthen instance-level representations. We therefore evaluate Qwen3-VL-Embedding-8B and our finetuned model on ILIAS~\citep{ilias2025}, INSTRE~\citep{wang2015instre}, PODS~\citep{sundaram2025personalized}, Paris6K~\citep{philbin2008lost}, Oxford5K~\citep{philbin2007object}, and Sculptures6K~\citep{arandjelovic2011smooth}, covering eight image--image settings and one text--image setting.


Table~\ref{tab:instance_retrieval} reports mAP across all settings. Our finetuned model improves over the pretrained model in every setting, with gains ranging from $+2.1$ to $+9.9$ mAP. To examine whether these gains reflect better discrimination among visually similar instances, we evaluate retrieval separately within ILIAS's fine-grained object categories, such as cakes or cards. Within each category, each image is used in turn as a query against the remaining images, with images depicting the same instance treated as relevant. Our model improves over the pretrained model in all $38$ fine-grained categories, with a mean mAP increase of $8.5$ points. Appendix~\ref{app:instance_retrieval} provides qualitative instance retrieval examples, as well as the full category-level analysis.

\vspace{-0.2cm}

\newcommand{\bodyfont}{\fontsize{10.95}{11.5}\selectfont}

\begin{table}[h]
\centering
\small
\caption{\textbf{Instance retrieval performance.}
Results report mean mAP (\%) across image--image and text--image benchmarks for Qwen3-VL-Embedding-8B and our finetuned model. For INSTRE, S and M denote the single- and multi-object subsets, respectively. Best results are shown in bold.
}
\label{tab:instance_retrieval}
\setlength{\tabcolsep}{5pt}
\renewcommand{\arraystretch}{1.4}

\resizebox{\linewidth}{!}{%
\begin{tabular}{l*{9}{c}}
\toprule
& \multicolumn{8}{c}{\textbf{I$\to$I}}
& \multicolumn{1}{c}{\textbf{T$\to$I}} \\
\cmidrule(lr){2-9}
\cmidrule(lr){10-10}

\textbf{\bodyfont Model}
& \textbf{ILIAS}
& \textbf{INSTRE S$\to$S}
& \textbf{INSTRE M$\to$S}
& \textbf{INSTRE M$\to$S+M}
& \textbf{PODS}
& \textbf{Paris6K}
& \textbf{Oxford5K}
& \textbf{Sculptures6K}
& \textbf{ILIAS} \\
\midrule

{\bodyfont Qwen3-VL-Emb-8B}
& {\bodyfont $58.5$}
& {\bodyfont $92.8$}
& {\bodyfont $81.3$}
& {\bodyfont $75.6$}
& {\bodyfont $85.7$}
& {\bodyfont $86.8$}
& {\bodyfont $61.3$}
& {\bodyfont $50.8$}
& {\bodyfont $52.3$} \\

{\bodyfont \textbf{Ours-8B}}
& {\bodyfont $\mathbf{68.4}$}
& {\bodyfont $\mathbf{94.9}$}
& {\bodyfont $\mathbf{86.7}$}
& {\bodyfont $\mathbf{79.0}$}
& {\bodyfont $\mathbf{88.1}$}
& {\bodyfont $\mathbf{88.9}$}
& {\bodyfont $\mathbf{64.3}$}
& {\bodyfont $\mathbf{59.6}$}
& {\bodyfont $\mathbf{60.8}$} \\

\bottomrule
\end{tabular}}
\vspace{-0.1cm}
\end{table}

\section{Discussion}
\label{discussion}
\vspace{-0.2cm}

Our results reveal a gap between broad retrieval capability and precise attribute--object binding. We introduce SABER to characterize these failures and show that targeted text-only finetuning improves binding across modality pairings, and that these gains transfer to independent benchmarks, negation, and instance retrieval, while largely preserving general multimodal performance. 
In Appendix~\ref{app:att_binding_beyond_swaps}, we show these gains extend beyond attribute swaps to a partial-description formulation. Appendix~\ref{app:instructions} shows that supplying the embedding model with binding-specific instructions helps but does not match the gains from our finetuning. A key limitation is that SABER is synthetic, and generated images may contain unintended variation in appearance, layout, object size, or orientation beyond the intended binding changes. Taken together, our findings suggest that compositional precision is a distinct dimension of embedding quality that warrants explicit evaluation and optimization.

\subsection*{AI use statement}
\label{sec_ai_use}

Generative AI tools were used for data generation and in several experimental
components of this work. Gemma 4 was used to generate synthetic text examples
for our finetuning dataset, while GPT-Image-2 was used to generate the synthetic SABER images
from author-defined object and attribute sets. Multiple vision--language models
were also used to propose and refine candidate instructions for the
instruction-sensitivity experiment; the authors reviewed and edited these
instructions, selected the final set, and conducted all evaluations.

Generative AI tools were additionally used to support code implementation,
figure design, and the drafting and editing of portions of the manuscript. AI tools were also used to assist with literature search and the identification of related work; all referenced works were independently verified and read by the authors. All AI-assisted code, figures, experimental materials, and manuscript text were
reviewed and approved by the authors.

These tools did not independently formulate the research questions, design the
methodology, interpret the results, or determine the conclusions. The authors
take full responsibility for all aspects of this work.

\subsection*{Ethics statement}
\label{sec:ethics}
This work does not involve the collection of new human-subject data and uses only existing research benchmarks and synthetically generated data. Some third-party benchmarks contain publicly sourced images, but we do not use identity information or analyze personal attributes. Attribute-binding performance should not be interpreted as evidence of broader fairness, safety, or suitability for deployment. Our results may inherit biases and limitations from the pretrained embedding models, generative models, and datasets used in this work. Licenses and applicable usage terms for third-party datasets and models are reported in Table~\ref{tab:licenses} in the Appendix.

\subsection*{Reproducibility statement}
\label{sec:repro}
We provide detailed descriptions of the SABER benchmark construction, textual finetuning data generation, training procedure, and evaluation protocols in the main paper and appendices. We will release SABER, our textual finetuning dataset, and the weights of our finetuned models to facilitate reproducibility and support future research on compositionality and attribute binding.

\subsubsection*{Acknowledgments}
We thank Stephanie Fu for helpful discussions throughout the project, and Narek Tumanyan for providing valuable comments on earlier versions of the paper.


\bibliography{iclr2027_conference}
\bibliographystyle{iclr2027_conference}

\clearpage

\appendix

\section{SABER Benchmark Details}
\label{app:dataset-details}

This section provides additional details on SABER's attribute sets, candidate construction, textual realization, image synthesis, and benchmark subsets.

\subsection{Building Blocks}
\label{sec:dataset-blocks}

The benchmark is generated from two lists: a set of $232$ everyday objects (e.g., toothbrush, teapot, kayak) and a set of $50$ visual attributes (e.g., red, glowing, sandy). The attributes are grouped into $6$ semantic categories, summarized in Table~\ref{tab:attributes}. In addition to a semantic category, each attribute is assigned a type that determines how it is realized during generation:

\begin{itemize}
    \item \textbf{\texttt{standalone}} ($27$): a fixed appearance with no free slot, such as \textit{camouflage} or \textit{transparent}. Some attributes use a fixed generation form; for example, \textit{soap-bubble} is expressed as \textit{iridescent soap-bubble}.
    \item \textbf{\texttt{single-color}} ($4$): a template with one color slot, e.g., \textit{glowing} $\rightarrow$ \textit{\{color\} glowing}.
    \item \textbf{\texttt{two-color}} ($6$): a template with two distinct color slots, e.g., \textit{striped} $\rightarrow$ \textit{\{color$_1$\} and \{color$_2$\} striped}.
    \item \textbf{\texttt{fixed-variant}} ($13$): a template instantiated from a small closed set of variants, e.g., \textit{wooden} $\rightarrow$ \textit{\{light-brown, dark-brown\} wooden}.
\end{itemize}

Color slots are filled from a palette of nine basic colors: red, orange, yellow, green, blue, purple, pink, white, and black. For two-color attributes, the two sampled colors are always distinct.

\begin{table}[h]
\centering
\small
\caption{\textbf{Number of SABER attributes in each semantic category.}}
\label{tab:attributes}
\begin{tabular}{lc}
\toprule
Attribute category & \# \\
\midrule
Basic color  & $9$  \\
Other color  & $2$  \\
Metallic     & $3$  \\
Material     & $19$ \\
Pattern      & $15$ \\
Effect       & $2$  \\
\midrule
Total        & $50$ \\
\bottomrule
\end{tabular}
\end{table}

\subsection{Example Structure}
\label{sec:dataset-structure}

To construct a multiple-choice question (MCQ) consisting of one query and four candidate answers, each with $N$ attribute--object bindings, we sample $N$ distinct objects and $N$ distinct attributes and pair them to form the query. We then derive four candidates, according to the
following rules:
\begin{itemize}
    \item \texttt{\textbf{candidate 1 (binding-preserving):}} preserves the
    query bindings while reordering the attribute--object pairs.

    \item \texttt{\textbf{candidate 2 (attribute-swap):}} keeps the objects in
    the same order as candidate~1 but permutes the attributes across them,
    producing incorrect attribute--object bindings.

    \item \texttt{\textbf{candidate 3 (object-swap):}} keeps the attributes in
    the same order as candidate~1 but permutes the objects across them,
    producing another incorrect binding assignment.

    \item \texttt{\textbf{candidate 4 (joint-swap):}} permutes both the objects
    and the attributes, producing another incorrect binding assignment.
\end{itemize}

For $N \geq 3$, the four candidates are required to have pairwise distinct binding assignments; any permutation that duplicates an existing candidate is rejected and resampled. For $N = 2$, where only two one-to-one assignments are possible, candidate~4 instead preserves one query binding and replaces either the object
or the attribute in the other binding with a new element not present in the query. For visualization purposes, the candidates are shown in shuffled order in Figures~\ref{fig:teaser} and~\ref{fig:dataset_a} of the main paper.

\subsection{Dual Textual Realization}
\label{sec:dataset-text}

Each query and candidate is expressed as a conjunction of its attribute--object pairs, such as
\textit{``A striped cube and a transparent kite.''} Indefinite articles
(\textit{a}/\textit{an}) are selected automatically based on pronunciation.
For each query and candidate, we produce two parallel textual forms:

\begin{itemize}
    \item the \textit{text} form, which uses the basic attribute forms (e.g., \textit{a striped cube}) and constitutes the benchmark's text modality;
    \item the \textit{generation} form, which uses the fully instantiated appearance (e.g., \textit{a white and orange striped cube}) and is used to construct the text-to-image prompt.
\end{itemize}

Within each example, every attribute is instantiated once and reused across the query and all candidates. For instance, if \textit{striped} is realized as \textit{white and orange striped}, that same realization is used wherever \textit{striped} appears. This keeps each attribute visually consistent across all images and reduces appearance variation unrelated to binding.

\subsection{Image Synthesis}
\label{sec:dataset-images}

Images are synthesized from the generation-form captions of the query and each candidate using GPT-Image-2 at $1024 \times 1024$ resolution and medium quality. The rendering prompt requests a CLEVR-style 3D scene on a neutral gray background in which the listed objects are clearly separated and scattered rather than laid out in a row, with each object appearing exactly once. A central requirement of the prompt is \textit{binding locality}: every attribute must apply only to its assigned object. In particular, the prompt specifies that colors, materials, patterns, textures, and visual effects must not transfer, blend, or bleed across objects. To reduce positional shortcuts, each rendered image is horizontally flipped with probability $0.5$. Finally, all generated images are manually screened to verify that the depicted attribute--object bindings match their associated captions.

\subsection{Benchmark Subsets}
\label{sec:dataset_subsets}

For each object count $N \in \{2,\ldots,10\}$, SABER contains $200$ MCQ examples, evenly split between independently sampled examples and nested chains. The independently sampled examples provide broader coverage of object, attribute, and binding combinations. In contrast, each nested chain starts with a two-object example and adds one new attribute--object binding at each subsequent object count, while reshuffling the order of the pairs. Thus, later queries preserve all bindings from earlier queries while introducing additional objects. Their binding-preserving candidates therefore serve as additional positives for earlier queries in large-gallery retrieval.

\section{Text-Only Finetuning for Attribute Binding}
\label{app:finetuning-dataset}

This section provides additional details on the construction of our text-only finetuning dataset, including its seven construction types, diversity controls, and validation procedure, as well as the training split, optimization setup, and evaluation of generalization to held-out binding structures.

\subsection{Generation Setup}

We generate the finetuning dataset using Gemma~4~31B-IT~\citep{gemmateam2026gemma4}.
For each example, the model selects the objects and attributes and writes a
contextualized positive caption expressing the intended attribute--object
bindings. From this structured output, we construct the query and two hard
negatives. The query lists the attribute--object pairs in
randomized order, while the negatives are formed by permuting either the
attributes or the objects in the positive caption. This short-query, longer-candidate format reflects practical retrieval, where a concise search query is matched against a gallery of items that may contain the queried content alongside additional information. Thus, the positive and both
negatives contain exactly the same words and differ only in their
attribute--object bindings. Each example is then automatically validated, with
failures discarded and regenerated.

The final dataset contains $100{,}000$ examples distributed across the seven
construction types shown in Table~\ref{tab:dataset-composition}.

\begin{table}[h]
\centering
\small
\caption{\textbf{Number of finetuning examples by construction type.}}
\label{tab:dataset-composition}
\begin{tabular}{lc}
\toprule
Construction type & \# \\
\midrule
pair\_swap                 & $20{,}000$ \\
attr\_bundle               & $20{,}000$ \\
predicative                & $20{,}000$ \\
triple\_bind\_cycle        & $10{,}048$ \\
triple\_bind\_swap         & $9{,}952$ \\
deferred\_bind\_anaphoric  & $9{,}417$ \\
deferred\_bind\_respective & $10{,}583$ \\
\midrule
Total                      & $100{,}000$ \\
\bottomrule
\end{tabular}
\end{table}

\subsection{Construction types}
\label{app:construction-types}
Each example belongs to one of seven construction types, which vary the number of attribute–object pairs and the sentence structure that expresses their bindings.
Table~\ref{tab:finetuning-dataset-examples} shows one example of each type:

\begin{itemize}
\item \textbf{\texttt{pair\_swap}}: two objects, each directly preceded by its
attribute. The attribute negative exchanges the two attributes, and the object
negative exchanges the two nouns.

\item \textbf{\texttt{triple\_bind\_cycle}}: three objects. The attribute
negative rotates the three attributes by one position, and the object negative
does the same with the nouns, so all three bindings are wrong in each negative.

\item \textbf{\texttt{triple\_bind\_swap}}: three objects, but only two pairs
are transposed. One of the three pairs, chosen at random, remains correctly
bound in both negatives, so the model must locate \emph{which} bindings
changed rather than simply detect that something changed.

\item \textbf{\texttt{attr\_bundle}}: two objects, each carrying two attributes
from different semantic slots (e.g., color and material, or age and
condition), which prevents redundant pairs such as \emph{``red crimson.''} The
attribute negative exchanges the attributes of only \emph{one} slot, so it
differs from the positive by a single word per object; the object negative
exchanges the nouns.

\item \textbf{\texttt{predicative}}: two objects whose attributes are detached
from their nouns through a copula, change-of-state verb, or relative clause
(e.g., \emph{``the book was dated''}). The adjacent phrase
``$\langle$attribute$\rangle$ $\langle$object$\rangle$'' may not appear
anywhere in the caption.

\item \textbf{\texttt{deferred\_bind\_anaphoric}}: both nouns are introduced
before either attribute appears, and the attributes are then assigned by
back-reference (e.g., \emph{``the former \ldots\ the latter''} or \emph{``the
first \ldots\ the second''}). In half of the examples the second object is
described first, so the reference must actually be resolved. Ambiguous
references such as a bare \emph{``it''} or \emph{``they''} are rejected.

\item \textbf{\texttt{deferred\_bind\_respective}}: both nouns are introduced
before either attribute appears, and the attributes are then assigned jointly
in noun order (e.g., \emph{``\ldots\ were new and neat respectively''}), so the
binding is carried entirely by word order.
\end{itemize}

In the \textbf{\texttt{predicative}} and \textbf{\texttt{deferred\_bind}} types, attributes are not adjacent to their objects, so the correct bindings must be recovered from sentence structure rather than local word co-occurrence.

\begin{table*}[t]
\centering
\small
\caption{\textbf{Examples of the seven finetuning construction types.}
Each block shows a query, positive caption, and two hard negatives formed by
permuting attributes or objects. Bold marks words moved relative to the positive.}
\label{tab:finetuning-dataset-examples}
\renewcommand{\arraystretch}{1.10}
\begin{tabularx}{\textwidth}{lX}
\toprule
\multicolumn{2}{l}{\textbf{\texttt{pair\_swap}}} \\
Query        & wrinkled tunic and cotton shroud \\
Positive     & ``Frost is coming,'' he noted, touching the cotton shroud and wrinkled tunic. \\
Neg.\ (attr.)& ``Frost is coming,'' he noted, touching the \textbf{wrinkled} shroud and \textbf{cotton} tunic. \\
Neg.\ (obj.) & ``Frost is coming,'' he noted, touching the cotton \textbf{tunic} and wrinkled \textbf{shroud}. \\
\midrule
\multicolumn{2}{l}{\textbf{\texttt{triple\_bind\_cycle}}} \\
Query        & plain chair, decorative lamp and metallic table \\
Positive     & ``It's eerie,'' says the mayor, beside a plain chair, metallic table and decorative lamp. \\
Neg.\ (attr.)& ``It's eerie,'' says the mayor, beside a \textbf{decorative} chair, \textbf{plain} table and \textbf{metallic} lamp. \\
Neg.\ (obj.) & ``It's eerie,'' says the mayor, beside a plain \textbf{lamp}, metallic \textbf{chair} and decorative \textbf{table}. \\
\midrule
\multicolumn{2}{l}{\textbf{\texttt{triple\_bind\_swap}}} \\
Query        & shared canvas, inexpensive palette and simple easel \\
Positive     & ``What a mess!'' cried he, seeing the shared canvas, simple easel and inexpensive palette. \\
Neg.\ (attr.)& ``What a mess!'' cried he, seeing the shared canvas, \textbf{inexpensive} easel and \textbf{simple} palette. \\
Neg.\ (obj.) & ``What a mess!'' cried he, seeing the shared canvas, simple \textbf{palette} and inexpensive \textbf{easel}. \\
\midrule
\multicolumn{2}{l}{\textbf{\texttt{attr\_bundle}}} \\
Query        & inviting delicate glass and ugly ridged plate \\
Positive     & Sipping tea, I used an inviting delicate glass and ugly ridged plate. \\
Neg.\ (attr.)& Sipping tea, I used an \textbf{ugly} delicate glass and \textbf{inviting} ridged plate. \\
Neg.\ (obj.) & Sipping tea, I used an inviting delicate \textbf{plate} and ugly ridged \textbf{glass}. \\
\midrule
\multicolumn{2}{l}{\textbf{\texttt{predicative}}} \\
Query        & torn page and dated book \\
Positive     & Suddenly, the book was dated and the page was torn per report. \\
Neg.\ (attr.)& Suddenly, the book was \textbf{torn} and the page was \textbf{dated} per report. \\
Neg.\ (obj.) & Suddenly, the \textbf{page} was dated and the \textbf{book} was torn per report. \\
\midrule
\multicolumn{2}{l}{\textbf{\texttt{deferred\_bind\_anaphoric}}} \\
Query        & hard umbrella and greenish awning \\
Positive     & Under the sun, an awning stood by an umbrella. The first was greenish, and the second felt hard. \\
Neg.\ (attr.)& Under the sun, an awning stood by an umbrella. The first was \textbf{hard}, and the second felt \textbf{greenish}. \\
Neg.\ (obj.) & Under the sun, an \textbf{umbrella} stood by an \textbf{awning}. The first was greenish, and the second felt hard. \\
\midrule
\multicolumn{2}{l}{\textbf{\texttt{deferred\_bind\_respective}}} \\
Query        & neat fence and new bench \\
Positive     & The bench by the path and the fence near the pinkish petal were new and neat respectively. \\
Neg.\ (attr.)& The bench by the path and the fence near the pinkish petal were \textbf{neat} and \textbf{new} respectively. \\
Neg.\ (obj.) & The \textbf{fence} by the path and the \textbf{bench} near the pinkish petal were new and neat respectively. \\
\bottomrule
\end{tabularx}
\end{table*}

\subsection{Diversity controls}
\label{app:diversity}
Repeatedly sampling a language model without additional constraints tends to
produce a narrow range of vocabulary, styles, and sentence structures. To
increase diversity, we therefore sample several properties independently for
each example and specify them in the prompt:

\begin{itemize}
\item \textbf{Context.} Each prompt samples from $1{,}000$ seed scenes, with
optional time-of-day and weather modifiers, yielding $81{,}000$ distinct
contexts. It additionally samples from $32$ writing registers (e.g., news
caption, field notebook, auction listing, maintenance log) and $16$
sentence-opener styles.

\item \textbf{Length.} Each example is assigned one of four target length bands
($10$--$15$, $16$--$22$, $23$--$30$, or $31$--$42$ words), shifted upward by
$2$--$4$ words for the three-object constructions. The target length band is
chosen so that it is long enough to accommodate the assigned binding-distance
range; the \textbf{\texttt{deferred\_bind}} constructions therefore use only
the three longer bands. The target band is specified in the prompt and enforced
during validation, allowing one word below and four words above the band.

\item \textbf{Binding distance.} Each example is assigned a target range for
the distance over which an attribute--object binding must be resolved,
preventing the relevant words from always appearing close together. For
constructions in which attributes directly precede their objects, this is the
minimum number of words separating consecutive core phrases (windows from
$0$--$2$ up to $14$--$26$ words for two objects, and up to $7$--$13$ for
three, where the minimum is taken over both gaps); for
\textbf{\texttt{predicative}}, it is the smaller of the two distances between an
object and its attribute; and for \textbf{\texttt{deferred\_bind}}, it is the
span from the final object mention to the first attribute assignment. The
target range is specified in the prompt and enforced during validation.

\item \textbf{Distractors.} For \textbf{\texttt{pair\_swap}},
\textbf{\texttt{predicative}}, and the two
\textbf{\texttt{deferred\_bind}} construction types, the prompt asks for up
to three additional attribute--object pairs that are woven into the sentence
but do not appear in the query. The positive text therefore describes more
than the query asks about, so the model must locate the queried bindings among
unrelated ones rather than treat every attribute--object pair in the text as
relevant.

\item \textbf{Vocabulary.} Examples that duplicate the complete set of
attributes and objects from an earlier example of the same construction type are rejected. To prevent a small
set of words from dominating the dataset, each attribute and object is subject
to a running frequency cap over the examples accepted so far. The cap is the
larger of ten occurrences and a fixed fraction of accepted examples, with the
fraction scaled by the number of attributes per example: $1.5\%$, $2.25\%$,
and $3\%$ for constructions with two, three, and four attributes,
respectively. This keeps the allowance per attribute slot constant while
preventing frequent words from becoming overrepresented early in generation.
Each prompt also includes an avoid list containing the $100$ most recently used
and $40$ most frequent words, together with a hard ban list for words that have
already reached the cap.

\item \textbf{Attribute pool.} Frequency caps alone were insufficient to
diversify the attributes, because the model continued to favor a small set of
common adjectives even after they had reached the cap, causing many generated
examples to be rejected. We therefore also provide attribute suggestions in
the prompt. These are drawn from a curated pool of $772$ common adjectives
organized into $19$ families that apply broadly across object classes, including
color, texture, material, condition, size, age, and value. The prompt includes
the least-used attributes from each family that have not yet reached the
frequency cap, sampled per family so that no single family dominates the
suggestions. The pool serves only as a source of suggestions: attributes
outside the pool are also accepted if they pass validation.
\end{itemize}

\subsection{Validation}
\label{app:validation}
Each generated example passes a structural validation step before being added to
the dataset; failures are discarded, logged, and regenerated. In addition to
the length, binding-distance, and vocabulary constraints described in
Section~\ref{app:diversity}, validation checks that the example has the intended
binding structure.

Attributes are required to be single words, and all objects and attributes must
be distinct, each appearing exactly once in the query, positive, and both
negatives, so that all queried bindings are specified unambiguously. Each
negative must be a word-level permutation of the positive (treating \emph{a}
and \emph{an} as equivalent) that breaks every queried binding, introduces all
intended incorrect bindings, and remains distinct from both the positive and
the other negative. Article-agreement errors created by a permutation (e.g.,
\emph{``a envelope''}) are repaired automatically, and any remaining article
error not already present in the positive causes the example to be rejected.

Finally, validation enforces the structural constraints of each construction:
attributes directly precede their objects in \textbf{\texttt{pair\_swap}},
\textbf{\texttt{triple\_bind}}, and \textbf{\texttt{attr\_bundle}}; are never adjacent to their
objects in \textbf{\texttt{predicative}}; and in \textbf{\texttt{deferred\_bind}} are never
adjacent to their objects and appear only after both objects have been
introduced.

\subsection{Finetuning and Evaluation}
\label{app:finetuning-use}
We hold out both \textbf{\texttt{deferred\_bind}} construction types entirely and
randomly split the remaining $80{,}000$ examples into training, validation, and
test sets in an $80/10/10$ ratio, yielding $64{,}000$, $8{,}000$, and
$8{,}000$ examples, respectively. Each example yields two triplets, one per hard negative, for a total of $200{,}000$ triplets: $128{,}000$ for training, $16{,}000$ for validation, $16{,}000$ for testing, and $40{,}000$ held-out \textbf{\texttt{deferred\_bind}} triplets.

Each model is finetuned using LoRA~\citep{hu2022lora} adapters with rank
$r{=}4$, $\alpha{=}4$, and zero dropout, applied to the query, key, value,
and output projections in every decoder layer. The vision tower remains
frozen, resulting in $3.8$M trainable parameters for the 8B model. All inputs
use the model's default instruction.

Training runs for one epoch on a single B200 GPU using
AdamW~\citep{loshchilov2018decoupled}, with a constant learning rate of
$5{\times}10^{-5}$ and weight decay $0.01$. The objective is a hinge triplet
loss over cosine similarity with margin $\gamma{=}0.05$. After training, we
apply LiNeS~\citep{wang2025lines} post hoc to rescale the LoRA updates across
decoder layers using the default setting of \citet{wang2025lines}.

We evaluate attribute-binding performance using triplet accuracy, defined as the fraction of triplets for which the positive is closer to the query than the negative; chance performance is $50\%$. Table~\ref{tab:finetuning-heldout} reports this metric across all evaluated models on the in-distribution test set and the two held-out \textbf{\texttt{deferred\_bind}} constructions, which test generalization to binding structures never seen during training.

\begin{table}[h]
\centering
\caption{%
\textbf{Triplet accuracy on held-out finetuning examples.}
Accuracy (\%) on the in-distribution test set and the two held-out deferred-binding construction types. Both of our finetuned models outperform all baselines across all three evaluations. Best results are shown in bold.
}
\label{tab:finetuning-heldout}

\normalsize
\setlength{\tabcolsep}{4pt}
\renewcommand{\arraystretch}{1.10}

\resizebox{\linewidth}{!}{%
\begin{tabular}{@{}l*{10}{c}@{}}
\toprule
\textbf{Dataset}
& {\small\makecell[c]{CLIP\\B/32}}
& {\small\makecell[c]{LABCLIP\\B/32}}
& {\small\makecell[c]{TripletCLIP\\B/32}}
& {\small\makecell[c]{READ-CLIP\\B/32}}
& {\small\makecell[c]{PE\\CORE-G}}
& {\small\makecell[c]{LamRA-\\Ret}}
& {\small\makecell[c]{Qwen3-VL\\Emb-2B}}
& {\small\makecell[c]{Qwen3-VL\\Emb-8B}}
& {\small\bfseries Ours-2B}
& {\small\bfseries Ours-8B} \\
\midrule

Test set
& 62.9 & 62.2 & 57.6 & 65.0 & 65.4 & 56.2 & 57.1 & 60.4
& 84.0 & \textbf{93.6} \\

deferred\_bind\_anaphoric
& 50.4 & 50.1 & 50.9 & 51.8 & 51.1 & 51.2 & 52.7 & 53.8
& 59.4 & \textbf{72.9} \\

deferred\_bind\_respective
& 51.0 & 50.9 & 52.1 & 51.5 & 51.1 & 51.1 & 51.4 & 50.8
& 53.2 & \textbf{56.6} \\

\bottomrule
\end{tabular}%
}
\end{table}

Our finetuned models improve substantially on the held-out test set, with
Ours-8B reaching $93.6\%$ triplet accuracy compared with $60.4\%$ for the
pretrained Qwen3-VL-Embedding-8B. The gains also generalize to the two
\textbf{\texttt{deferred\_bind}} constructions that are never seen during
training: Ours-8B improves from $53.8\%$ to $72.9\%$ on
\textbf{\texttt{deferred\_bind\_anaphoric}} and from $50.8\%$ to $56.6\%$ on
\textbf{\texttt{deferred\_bind\_respective}}. This indicates that the learned
binding improvements are not limited to the specific sentence structures used
for finetuning.

\section{Ablations}
\label{app:ablations}

\subsection{Training Data}

Table~\ref{tab:data_ablation} first tests whether the binding gains in
Section~\ref{results} reflect a general compositional capability or merely
exposure to SABER's vocabulary. We retrain on three size-matched subsets of
the training pool ($30{,}900$ rows each): \emph{Qwen3-VL-Emb-8B-NoOverlap},
containing only examples with no SABER attributes or objects;
\emph{Qwen3-VL-Emb-8B-Random}, a uniform random sample preserving the pool's
natural lexical overlap with SABER; and
\emph{Qwen3-VL-Emb-8B-OverlapOnly}, containing only examples that mention at
least one SABER attribute or object.
Even without any vocabulary overlap, finetuning yields large gains across all
four SABER pairings, improving text--text accuracy from $53.7$ to $99.3$,
text--image from $34.9$ to $65.7$, image--text from $36.3$ to $71.8$, and
image--image from $84.9$ to $93.1$. Increasing lexical overlap with SABER
provides only modest additional gains. Relative to
\emph{Qwen3-VL-Emb-8B-NoOverlap}, \emph{Qwen3-VL-Emb-8B-Random} improves mean
SABER accuracy by $1.6$ points, while \emph{Qwen3-VL-Emb-8B-OverlapOnly}
improves it by $2.0$ points. MMEB-v3 performance remains nearly unchanged
across the three training subsets, with the overall score ranging from $55.2$
to $55.3$ and closely matching our finetuned 8B model. Overall, the results
indicate that the binding gains primarily reflect transferable binding
supervision, with SABER-specific lexical overlap contributing only modestly.

\begin{table}[h]
\centering
\small
\caption{
\textbf{Training-data ablation.}
SABER-MCQ and MMEB-v3 performance for lexical-overlap and query--candidate
length ablations of our 8B model. SABER-MCQ reports 4-way accuracy, while
MMEB-v3 reports Hit@1 for Image, Video, and Agent and NDCG@5 for VisDoc and
Text. Best results are shown in bold.
}
\label{tab:data_ablation}
\setlength{\tabcolsep}{4pt}
\renewcommand{\arraystretch}{1.15}
\setlength{\arrayrulewidth}{0.6pt}

\resizebox{\linewidth}{!}{%
\begin{tabular}{@{}lcccc|cccccc@{}}
\toprule
&
\multicolumn{4}{c}{\textbf{SABER-MCQ}}
& \multicolumn{6}{c}{\textbf{MMEB-v3}} \\
\cmidrule(lr){2-5}
\cmidrule(lr){6-11}

\textbf{Model}
& \textbf{T$\to$T}
& \textbf{T$\to$I}
& \textbf{I$\to$T}
& \textbf{I$\to$I}
& \textbf{Image}
& \textbf{Video}
& \textbf{VisDoc}
& \textbf{Text}
& \textbf{Agent}
& \textbf{All} \\
\midrule

Qwen3-VL-Emb-8B-NoOverlap
& $99.3$ & $65.7$ & $71.8$ & $93.1$
& $72.6$ & $58.8$ & $75.5$ & $43.1$ & $\mathbf{42.6}$ & $\mathbf{55.3}$ \\

Qwen3-VL-Emb-8B-Random
& $99.4$ & $68.7$ & $73.8$ & $94.3$
& $72.5$ & $58.8$ & $75.4$ & $43.2$ & $42.3$ & $\mathbf{55.3}$ \\

Qwen3-VL-Emb-8B-OverlapOnly
& $99.4$ & $69.8$ & $74.1$ & $94.7$
& $72.5$ & $58.8$ & $75.4$ & $43.3$ & $42.0$ & $55.2$ \\

Qwen3-VL-Emb-8B-MixedLengths
& $\mathbf{99.6}$ & $\mathbf{70.6}$ & $\mathbf{77.1}$ & $\mathbf{94.8}$
& $72.3$ & $58.8$ & $\mathbf{75.6}$ & $42.9$ & $\mathbf{42.6}$ & $55.2$ \\

\midrule

Qwen3-VL-Emb-8B
& $53.7$ & $34.9$ & $36.3$ & $84.9$
& $\mathbf{73.7}$ & $\mathbf{59.6}$ & $74.3$ & $\mathbf{43.7}$ & $39.1$ & $54.7$ \\

\textbf{Ours-8B}
& $99.3$ & $69.0$ & $74.1$ & $94.3$
& $72.5$ & $58.7$ & $75.4$ & $43.1$ & $42.3$ & $55.2$ \\

\bottomrule
\end{tabular}%
}
\end{table}

The final data ablation changes the structure of the training data rather than its
lexical overlap with SABER. Our finetuned 8B model is trained only on
short-query, long-candidate examples, where a short conjunction of
attribute--object pairs is matched against a contextualized positive sentence
and negatives formed by swapping attributes or objects. For
\emph{Qwen3-VL-Emb-8B-MixedLengths}, we augment this data with two additional
formats constructed from the same underlying examples: short queries with
short candidates and long queries with short candidates. The resulting
training set contains $192{,}000$ examples and exposes the same bindings across
all three query--candidate length configurations.
\emph{Qwen3-VL-Emb-8B-MixedLengths} improves over our finetuned 8B model
across all four SABER pairings, increasing mean accuracy by $1.4$ points.
MMEB-v3 performance, however, is essentially unchanged, with the overall score
remaining at $55.2$. Because the mixed-length training set includes a short-query, short-candidate format that more closely
matches SABER's templated query--candidate structure, we do not interpret
these gains as evidence of better general attribute binding. Instead, the
higher SABER scores may partly reflect closer alignment between the training
and evaluation formats.

\subsection{Effect of LiNeS}
\label{app:abl_lines_effect}

We next examine the effect of LiNeS~\citep{wang2025lines} on the trade-off between attribute binding and general multimodal embedding performance. We evaluate \emph{Qwen3-VL-Emb-8B-NoLiNeS} after finetuning but before applying LiNeS, with all other settings unchanged. As shown in Table~\ref{tab:reg_ablation}, it achieves higher SABER accuracy than our final model on all four modality pairings, improving mean accuracy by $9.4$ points. However, MMEB-v3 performance drops substantially, from $55.2$ to $47.8$, well below the pretrained model's $54.7$. Thus, LiNeS sacrifices some binding gains to substantially recover general multimodal embedding performance.

\begin{table}[h]
\centering
\small
\caption{
\textbf{LiNeS ablation.}
SABER-MCQ and MMEB-v3 performance for our 8B model before and after applying
LiNeS. SABER-MCQ reports 4-way accuracy, while MMEB-v3 reports Hit@1 for
Image, Video, and Agent and NDCG@5 for VisDoc and Text. Best results are shown in bold.
}
\label{tab:reg_ablation}
\setlength{\tabcolsep}{4pt}
\renewcommand{\arraystretch}{1.15}
\setlength{\arrayrulewidth}{0.6pt}

\resizebox{\linewidth}{!}{%
\begin{tabular}{@{}lcccc|cccccc@{}}
\toprule
&
\multicolumn{4}{c}{\textbf{SABER-MCQ}}
& \multicolumn{6}{c}{\textbf{MMEB-v3}} \\
\cmidrule(lr){2-5}
\cmidrule(lr){6-11}

\textbf{Model}
& \textbf{T$\to$T}
& \textbf{T$\to$I}
& \textbf{I$\to$T}
& \textbf{I$\to$I}
& \textbf{Image}
& \textbf{Video}
& \textbf{VisDoc}
& \textbf{Text}
& \textbf{Agent}
& \textbf{All} \\
\midrule

Qwen3-VL-Emb-8B-NoLiNeS
& $\mathbf{100.0}$ & $\mathbf{87.3}$ & $\mathbf{87.8}$ & $\mathbf{99.0}$
& $60.8$ & $51.2$ & $68.8$ & $34.6$ & $39.0$ & $47.8$ \\

\midrule

Qwen3-VL-Emb-8B
& $53.7$ & $34.9$ & $36.3$ & $84.9$
& $\mathbf{73.7}$ & $\mathbf{59.6}$ & $74.3$ & $\mathbf{43.7}$ & $39.1$ & $54.7$ \\

\textbf{Ours-8B}
& $99.3$ & $69.0$ & $74.1$ & $94.3$
& $72.5$ & $58.7$ & $\mathbf{75.4}$ & $43.1$ & $\mathbf{42.3}$ & $\mathbf{55.2}$ \\

\bottomrule
\end{tabular}%
}
\end{table}

\subsection{Attention-Based Binding Analysis and Direct Supervision}

To test whether finetuning changes how attributes and objects interact, we directly analyze text-side attention. Because causal attention flows only from later to earlier tokens, we measure each object--attribute connection in the available direction, from the later word to the earlier one. For words spanning multiple sub-tokens, we sum the attention from each sub-token of the later word to all sub-tokens of the earlier word, then average over the later-word sub-tokens and attention heads to obtain one value per layer and word pair. For each object, we define a binding margin as its mean attention to its own attribute words minus its mean attention to the attributes of the other objects in the same caption, using the same later-to-earlier convention. We average these object-level margins to obtain one margin per caption and layer; zero indicates no preference for the correct attributes. We compute this margin across all 36 text layers for Qwen3-VL-Embedding-8B, our finetuned model, and our finetuned model without LiNeS, on three text sets: $1{,}000$ short queries sampled from the held-out test split in proportion to the training distribution over construction types, their corresponding longer positive captions, and $1{,}000$ SABER queries spanning two to ten objects.

\begin{figure*}[h]
    \centering
    \includegraphics[width=\textwidth]{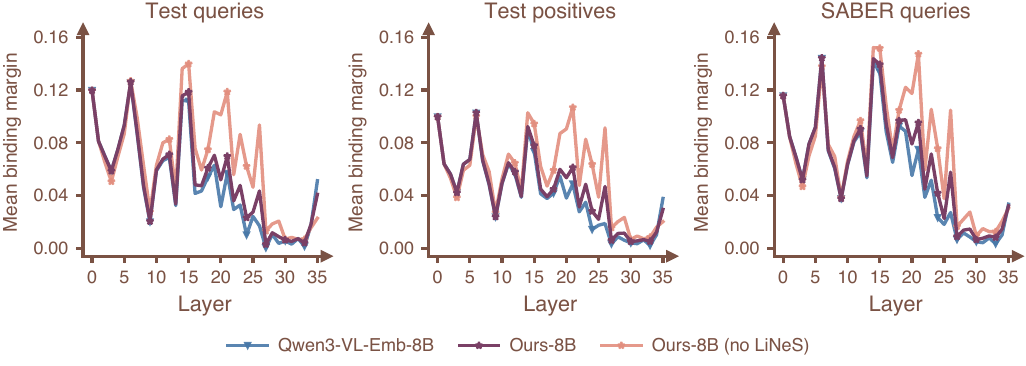}
    \caption{\textbf{Layer-wise binding margin.} Mean binding margin across the 36-layer text stack for test queries, their corresponding positive captions, and SABER queries. Finetuning preserves stronger binding in later layers, with a larger effect without LiNeS.}
    \label{fig:attn_by_layer}
\end{figure*}

Figure~\ref{fig:attn_by_layer} shows a consistent pattern across all three text sets. The base model exhibits positive binding margins in early layers, but this signal decays toward zero in the final third of the network. Finetuning leaves the first $\sim11$ layers largely unchanged, while strong differences emerge in layers 18--26, where the base margin collapses. Over this band, Ours-8B retains $1.3$--$1.4\times$ the base margin, and the variant without LiNeS retains $1.9$--$2.4\times$, across queries, positives, and SABER. This suggests that finetuning helps preserve existing attribute--object structure deeper into the network, where it can influence the pooled embedding. We therefore turn this observation into an attention loss, described next, applied to this same layer range.

For a text input $x$, let $\mathcal{O}(x)$ denote its object words and
$\mathcal{A}_{x}{(o)}$ the attribute words associated with object $o$ in input $x$.
Let $\overline{\mathcal{A}}_{x}{(o)}$ denote the attribute words associated with the
other objects in $x$. We write $e_x^\ell(u,v)$ for the head-averaged
attention between words $u$ and $v$ at layer $\ell$, measured in the
causally available direction from the later word to the earlier one and
aggregated over their sub-tokens.

We define the binding margin of object $o$ at layer $\ell$ as
\begin{equation}
m_x^\ell(o)
=
\frac{1}{|\mathcal{A}_x(o)|}
\sum_{a \in \mathcal{A}_x(o)} e_x^\ell(o,a)
-
\frac{1}{|\overline{\mathcal{A}}_x(o)|}
\sum_{a \in \overline{\mathcal{A}}_x(o)} e_x^\ell(o,a).
\end{equation}

Let $\mathcal{B}$ denote a batch of text queries,
$\mathcal{L}=\{18,\ldots,26\}$ the layers on which we apply the attention loss,
$[z]_+=\max(0,z)$ the positive-part function, and $\tau=0.1$ the target binding margin.
We penalize object--layer pairs whose margin falls below $\tau$:
\begin{equation}
\mathcal{L}_{\mathrm{attn}}
=
\frac{1}{|\mathcal{L}|\sum_{x\in\mathcal{B}}|\mathcal{O}(x)|}
\sum_{x\in\mathcal{B}}
\sum_{o\in\mathcal{O}(x)}
\sum_{\ell\in\mathcal{L}}
\bigl[\tau-m_x^\ell(o)\bigr]_+ .
\end{equation}

The total training objective combines the triplet and attention losses, with
$\lambda_{\mathrm{attn}}=0.1$ weighting the latter:
\begin{equation}
\mathcal{L}
=
\mathcal{L}_{\mathrm{triplet}}
+
\lambda_{\mathrm{attn}}\mathcal{L}_{\mathrm{attn}}.
\end{equation}

\begin{table}[h]
\centering
\small
\caption{
\textbf{Attention loss ablation.}
SABER-MCQ and MMEB-v3 performance for our 8B model trained with the attention loss, with and without LiNeS. SABER-MCQ reports 4-way
accuracy, while MMEB-v3 reports Hit@1 for Image, Video, and Agent and NDCG@5
for VisDoc and Text. Best results are shown in bold.
}
\label{tab:attn_ablation}
\setlength{\tabcolsep}{4pt}
\renewcommand{\arraystretch}{1.15}
\setlength{\arrayrulewidth}{0.6pt}

\resizebox{\linewidth}{!}{%
\begin{tabular}{@{}lcccc|cccccc@{}}
\toprule
&
\multicolumn{4}{c}{\textbf{SABER-MCQ}}
& \multicolumn{6}{c}{\textbf{MMEB-v3}} \\
\cmidrule(lr){2-5}
\cmidrule(lr){6-11}

\textbf{Model}
& \textbf{T$\to$T}
& \textbf{T$\to$I}
& \textbf{I$\to$T}
& \textbf{I$\to$I}
& \textbf{Image}
& \textbf{Video}
& \textbf{VisDoc}
& \textbf{Text}
& \textbf{Agent}
& \textbf{All} \\
\midrule

Qwen3-VL-Emb-8B-Attn-LiNeS
& $99.4$ & $70.0$ & $74.8$ & $94.2$
& $72.4$ & $58.6$ & $75.2$ & $42.8$ & $42.2$ & $55.0$ \\

Qwen3-VL-Emb-8B-Attn-NoLiNeS
& $\mathbf{100.0}$ & $\mathbf{86.4}$ & $\mathbf{87.2}$ & $\mathbf{99.2}$
& $59.5$ & $50.5$ & $66.3$ & $34.8$ & $39.2$ & $47.2$ \\

\midrule

Qwen3-VL-Emb-8B
& $53.7$ & $34.9$ & $36.3$ & $84.9$
& $\mathbf{73.7}$ & $\mathbf{59.6}$ & $74.3$ & $\mathbf{43.7}$ & $39.1$ & $54.7$ \\

\textbf{Ours-8B}
& $99.3$ & $69.0$ & $74.1$ & $94.3$
& $72.5$ & $58.7$ & $\mathbf{75.4}$ & $43.1$ & $\mathbf{42.3}$ & $\mathbf{55.2}$ \\

\bottomrule
\end{tabular}%
}
\end{table}

Training with the combined objective yields the results in
Table~\ref{tab:attn_ablation}. With LiNeS, adding the attention loss provides
only a modest improvement, increasing mean SABER accuracy by $0.4$ points over
Ours-8B while slightly reducing MMEB-v3 from $55.2$ to $55.0$. Without LiNeS,
the model reaches much higher SABER accuracy, but this behavior is already
observed when removing LiNeS without attention supervision
(Section~\ref{app:abl_lines_effect}). In fact, adding the attention loss to the setting without LiNeS
slightly decreases mean SABER accuracy from $93.5$ to $93.2$ and reduces
MMEB-v3 from $47.8$ to $47.2$. Overall, direct attention supervision provides
little additional benefit beyond our existing training objective.

\section{Generalization Across Multimodal Embedders}
\label{app:diff_backbones}
To test whether our finetuning strategy is specific to Qwen3-VL-Embedding or
generalizes to other multimodal embedders, we apply the same recipe to two
additional model families, UEmbed~\citep{uembed2026} and
WeMM-Embedding~\citep{wemm-embedding}, each at 2B and 9B parameters. Both are
built on Qwen3.5~\citep{qwen35}, whose hybrid decoder uses softmax attention only every fourth
layer, with linear-attention blocks in the remaining layers.

We retain the training setup from our main experiments, using the same data,
learning rate, and LiNeS. We apply rank-4 LoRA to the query, key, value, and
output projections of the full-attention layers and evaluate the checkpoint
after one epoch. Each model uses its official prompting and pooling scheme; for UEmbed, we use the dense embedding rather than its sparse representation. The unmodified
public checkpoint serves as the baseline for comparison with its finetuned
counterpart on SABER and MMEB-v3.

As shown in Table~\ref{tab:other_models}, binding-focused finetuning consistently improves SABER across both model families, model sizes, and all four modality pairings. The largest gains generally occur in text--text retrieval: UEmbed improves by $48.6$ and $44.3$ points at 2B and 9B, respectively, while WeMM-Embedding improves by $37.5$ and $31.9$ points. Importantly, these gains transfer beyond the text modality used for training. All four models also improve on both cross-modal settings and on image--image retrieval, with particularly large cross-modal gains for the 9B models. For example, WeMM-Embedding-9B improves from $30.6$ to $58.7$ in text--image retrieval and from $32.6$ to $64.4$ in image--text retrieval. This consistent transfer suggests that the effect is not specific to the Qwen3-VL-Embedding backbone and that text-only binding supervision can improve multimodal representations across different embedding architectures.

\begin{table}[h]
\centering
\small
\caption{
\textbf{Generalization to additional embedding backbones.}
Performance of UEmbed and WeMM-Embedding on SABER and MMEB-v3 before and after binding-focused finetuning. Finetuning improves all four SABER modality pairings across both model families and scales, while largely preserving MMEB-v3 performance. SABER-MCQ reports 4-way accuracy, while MMEB-v3 reports Hit@1 for Image, Video, and Agent and NDCG@5 for VisDoc and Text. Best results per pretrained--finetuned pair are shown in bold.}
\label{tab:other_models}
\setlength{\tabcolsep}{4pt}
\renewcommand{\arraystretch}{1.15}
\setlength{\arrayrulewidth}{0.6pt}

\resizebox{\linewidth}{!}{%
\begin{tabular}{@{}lcccc|cccccc@{}}
\toprule
&
\multicolumn{4}{c}{\textbf{SABER-MCQ}}
& \multicolumn{6}{c}{\textbf{MMEB-v3}} \\
\cmidrule(lr){2-5}
\cmidrule(lr){6-11}

\textbf{Model}
& \textbf{T$\to$T}
& \textbf{T$\to$I}
& \textbf{I$\to$T}
& \textbf{I$\to$I}
& \textbf{Image}
& \textbf{Video}
& \textbf{VisDoc}
& \textbf{Text}
& \textbf{Agent}
& \textbf{All} \\
\midrule

UEmbed-2B
& $33.2$ & $27.3$ & $24.7$ & $83.4$
& $\mathbf{63.7}$ & $\mathbf{50.8}$ & $\mathbf{71.5}$ & $\mathbf{35.4}$ & $\mathbf{35.9}$ & $\mathbf{48.2}$ \\

UEmbed-2B-Finetuned
& $\mathbf{81.8}$ & $\mathbf{39.2}$ & $\mathbf{39.8}$ & $\mathbf{88.3}$
& $62.2$ & $48.9$ & $67.8$ & $31.9$ & $34.2$ & $45.6$ \\
\midrule

UEmbed-9B
& $46.9$ & $36.4$ & $32.2$ & $88.7$
& $\mathbf{68.6}$ & $\mathbf{55.6}$ & $\mathbf{75.2}$ & $\mathbf{40.9}$ & $\mathbf{38.1}$ & $\mathbf{52.3}$ \\

UEmbed-9B-Finetuned
& $\mathbf{91.2}$ & $\mathbf{57.3}$ & $\mathbf{49.7}$ & $\mathbf{94.6}$
& $67.6$ & $55.1$ & $74.5$ & $40.4$ & $35.9$ & $51.3$ \\
\midrule

WeMM-Embedding-2B
& $51.4$ & $29.8$ & $28.9$ & $59.6$
& $\mathbf{78.0}$ & $\mathbf{65.2}$ & $\mathbf{78.4}$ & $\mathbf{44.2}$ & $44.8$ & $\mathbf{58.4}$ \\

WeMM-Embedding-2B-Finetuned
& $\mathbf{88.9}$ & $\mathbf{38.9}$ & $\mathbf{41.2}$ & $\mathbf{73.3}$
& $77.1$ & $64.3$ & $77.6$ & $43.9$ & $\mathbf{45.9}$ & $58.2$ \\
\midrule

WeMM-Embedding-9B
& $65.4$ & $30.6$ & $32.6$ & $63.7$
& $\mathbf{80.2}$ & $\mathbf{68.5}$ & $\mathbf{81.1}$ & $\mathbf{46.9}$ & $\mathbf{50.1}$ & $\mathbf{61.7}$ \\

WeMM-Embedding-9B-Finetuned
& $\mathbf{97.3}$ & $\mathbf{58.7}$ & $\mathbf{64.4}$ & $\mathbf{78.6}$
& $79.7$ & $67.9$ & $80.1$ & $46.7$ & $49.4$ & $61.1$ \\

\bottomrule
\end{tabular}%
}
\end{table}

These gains come with relatively modest changes in general-purpose retrieval performance. MMEB-v3 decreases by $1.0$ point for UEmbed-9B and by only $0.2$ and $0.6$ points for WeMM-Embedding-2B and 9B, respectively, despite their large SABER improvements. UEmbed-2B shows a larger $2.6$-point decrease, indicating a less favorable binding--generalization trade-off at this scale. Overall, however, the same lightweight finetuning recipe substantially improves attribute binding across additional backbones while largely retaining their broader retrieval capabilities.

\section{Comparing Text and Image Finetuning for Attribute Binding}
\label{app:text_vs_images}

Our main approach uses text-only supervision because text makes
attribute--object bindings explicit and easy to manipulate, enabling precisely
controlled hard negatives. However, this high degree of control may also make
text supervision artificially clean. Images inevitably introduce variation in
appearance, pose, lighting, size, spatial arrangement, and other visual factors.
Although such variation makes the learning problem harder, it may also provide
useful regularization, forcing the model to learn binding representations that
are robust to nuisance variation rather than relying on the clean structure of
text. We therefore ask whether text or image supervision is more effective for
learning attribute binding. To study this, we split SABER into a $70/10/20$ train/validation/test split
and finetune on increasing amounts of either text or image supervision,
corresponding to $10\%$, $20\%$, $35\%$, and $70\%$ of the full benchmark. Results are shown in Table~\ref{tab:text_vs_images}.

\begin{table}[h]
\centering
\small
\caption{
\textbf{Text vs.\ image finetuning on SABER.}
Performance of Qwen3-VL-Embedding-8B after finetuning with increasing amounts of SABER data using either text--text or image--image supervision. Text supervision yields stronger overall SABER performance and consistently better cross-modal transfer, while image supervision performs best on the matched image--image setting. SABER--MCQ reports 4-way accuracy on the fixed $360$-example test split, with all models evaluated at epoch $40$. MMEB-v3 reports Hit@1 for Image, Video, and Agent tasks and NDCG@5 for VisDoc and Text tasks. Best results are shown in bold.
}
\label{tab:text_vs_images}
\setlength{\tabcolsep}{4pt}
\renewcommand{\arraystretch}{1.15}
\setlength{\arrayrulewidth}{0.6pt}

\resizebox{\linewidth}{!}{%
\begin{tabular}{@{}cccccc|cccccc@{}}
\toprule
&
&
\multicolumn{4}{c}{\textbf{\shortstack{SABER-MCQ\\360 Test Examples (20\%)}}}& \multicolumn{6}{c}{\textbf{MMEB-v3}} \\
\cmidrule(lr){3-6}
\cmidrule(lr){7-12}
\noalign{\vskip -2.2ex}

\textbf{\shortstack{Finetuning\\Modalities}}
&\textbf{\shortstack{SABER \\Finetuning Data}}
& \textbf{T$\to$T}
& \textbf{T$\to$I}
& \textbf{I$\to$T}
& \textbf{I$\to$I}
& \textbf{Image}
& \textbf{Video}
& \textbf{VisDoc}
& \textbf{Text}
& \textbf{Agent}
& \textbf{All} \\
\midrule

\multirow{4}{*}{T$\to$T}
& 180 (10\%)
& $87.2$ & $40.6$ & $46.7$ & $87.5$
& $73.8$ & $59.8$ & $74.8$ & $43.7$ & $39.3$ & $54.9$ \\

& 360 (20\%)
& $95.6$ & $52.5$ & $55.6$ & $90.0$
& $\mathbf{73.9}$ & $\mathbf{59.9}$ & $74.9$ & $43.7$ & $39.5$ & $55.0$ \\

& 630 (35\%)
& $96.9$ & $54.7$ & $59.4$ & $90.6$
& $73.8$ & $\mathbf{59.9}$ & $75.0$ & $\mathbf{43.8}$ & $\mathbf{39.7}$ & $\mathbf{55.1}$ \\

& 1,260 (70\%)
& $\mathbf{97.8}$ & $\mathbf{55.6}$ & $\mathbf{60.3}$ & $90.8$
& $73.8$ & $59.8$ & $75.1$ & $43.7$ & $39.6$ & $55.0$ \\

\midrule

\multirow{4}{*}{I$\to$I}
& 180 (10\%)
& $56.1$ & $32.8$ & $39.4$ & $95.3$
& $\mathbf{73.9}$ & $59.6$ & $74.6$ & $43.7$ & $39.2$ & $54.8$ \\

& 360 (20\%)
& $61.1$ & $36.7$ & $44.4$ & $96.9$
& $\mathbf{73.9}$ & $59.7$ & $75.0$ & $43.5$ & $39.5$ & $54.9$ \\

& 630 (35\%)
& $66.1$ & $38.6$ & $51.9$ & $97.8$
& $\mathbf{73.9}$ & $59.5$ & $75.0$ & $43.4$ & $39.6$ & $54.9$ \\

& 1,260 (70\%)
& $67.2$ & $42.5$ & $52.8$ & $\mathbf{99.2}$
& $73.8$ & $59.6$ & $\mathbf{75.2}$ & $43.5$ & $39.5$ & $54.9$ \\

\bottomrule
\end{tabular}
}
\end{table}

Across both supervision modalities, SABER performance improves as more finetuning data is used. As expected, each modality performs best in its matched evaluation setting: text--text finetuning yields substantially higher text--text accuracy, whereas image--image finetuning is consistently stronger on image--image retrieval. More interestingly, within this controlled SABER setting, text supervision transfers better to the two cross-modal evaluations. At every training-set size, text--text finetuning outperforms image--image finetuning on both text--image and image--text retrieval. With the full training split, the gap is $13.1$ points in text--image retrieval and $7.5$ points in image--text retrieval.

MMEB-v3 performance remains largely stable across both supervision modalities and training-set sizes, with the overall score varying only from $54.8$ to $55.1$, possibly reflecting the relatively small amount of finetuning data used here. While the SABER results suggest that text supervision may provide a more effective signal for cross-modal binding transfer, this comparison is limited to a relatively small benchmark with a single family of synthetic images and templated texts, and should not be taken as evidence that text supervision is generally superior to image supervision.

\section{Evaluation and Analysis on Instance Retrieval}
\label{app:instance_retrieval}

\subsection{Datasets and Evaluation Protocols}

We evaluate on six instance-retrieval benchmarks spanning landmarks, sculptures, household objects, and diverse object instances. Oxford5K~\citep{philbin2007object} and Paris6K~\citep{philbin2008lost} evaluate landmark retrieval using cropped query regions. Images labeled \textit{good} or \textit{ok} are treated as relevant, while \textit{junk} images are ignored during evaluation. Sculptures6K~\citep{arandjelovic2011smooth} similarly uses cropped queries, with benchmark-specific relevance annotations. PODS~\citep{sundaram2025personalized} contains household objects photographed under several distribution shifts and evaluates retrieval across reference views of each object. ILIAS~\citep{ilias2025} contains diverse object instances and supports both image--image and text--image retrieval over a shared image database. INSTRE~\citep{wang2015instre} contains single-object images (INSTRE-S) and images depicting pairs of objects (INSTRE-M). Following the original protocol, we evaluate queries from INSTRE-S against INSTRE-S, and queries from INSTRE-M against either INSTRE-S or the combined INSTRE-S+M gallery. In the combined setting, relevance depends on whether a retrieved image contains one or both query instances, with images containing both receiving higher relevance.

We evaluate Qwen3-VL-Embedding-8B and our finetuned model under identical conditions on each benchmark, following the benchmark-specific protocols described above. Images are embedded in full unless a benchmark provides a ground-truth query crop, in which case we use the specified crop. We do not include auxiliary distractor sets. Accordingly, comparisons between the pretrained and finetuned models are directly controlled, while absolute performance may differ from published results that use additional distractors.

\begin{figure*}[h]
    \centering
    \includegraphics[width=\textwidth]{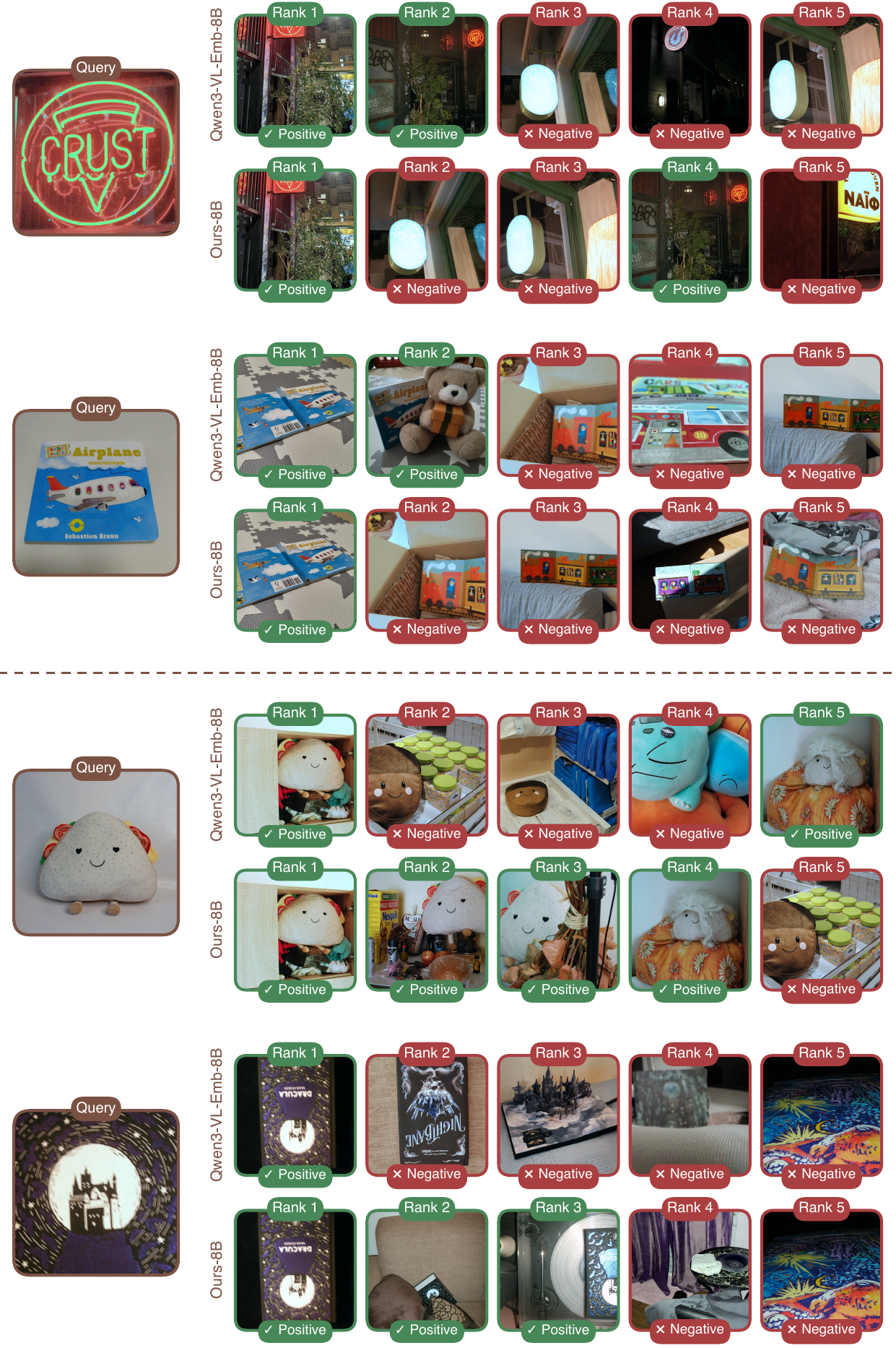}
\caption{\textbf{Qualitative image--image retrieval on ILIAS.}
Top-5 retrievals for two queries favoring Qwen3-VL-Embedding-8B (top) and two favoring Ours-8B (bottom). Green positives denote images of the same instance as the query, while red negatives denote other instances.}
    \label{fig:instance_retrieval}
\end{figure*}

Quantitative results across all benchmarks and evaluation settings are reported in Table~\ref{tab:instance_retrieval} of the main paper. Figure~\ref{fig:instance_retrieval} complements these results with qualitative image--image retrieval examples on ILIAS. We show queries for which either the pretrained or finetuned model has a large relative advantage, together with their top-5 retrieved images. Across all $1{,}232$ ILIAS image queries, the finetuned model performs better on $923$, the pretrained model on $108$, and the two models tie on $201$.

\subsection{Why Might Binding-Focused Finetuning Improve Instance Retrieval?}
\label{app:instance_retrieval_analysis}

To examine the instance-retrieval gains more closely, we use ILIAS's hierarchical taxonomy to evaluate retrieval within fine-grained object classes. We consider its $38$ named fine-grained classes, spanning $491$ object instances across seven top-level categories. We exclude leaves labeled \emph{other}, while the Media category contains no named class at this level. Retrieval is evaluated independently within each class using the protocol described in the main paper.

\begin{figure*}[h]
    \centering
    \includegraphics[width=\textwidth]{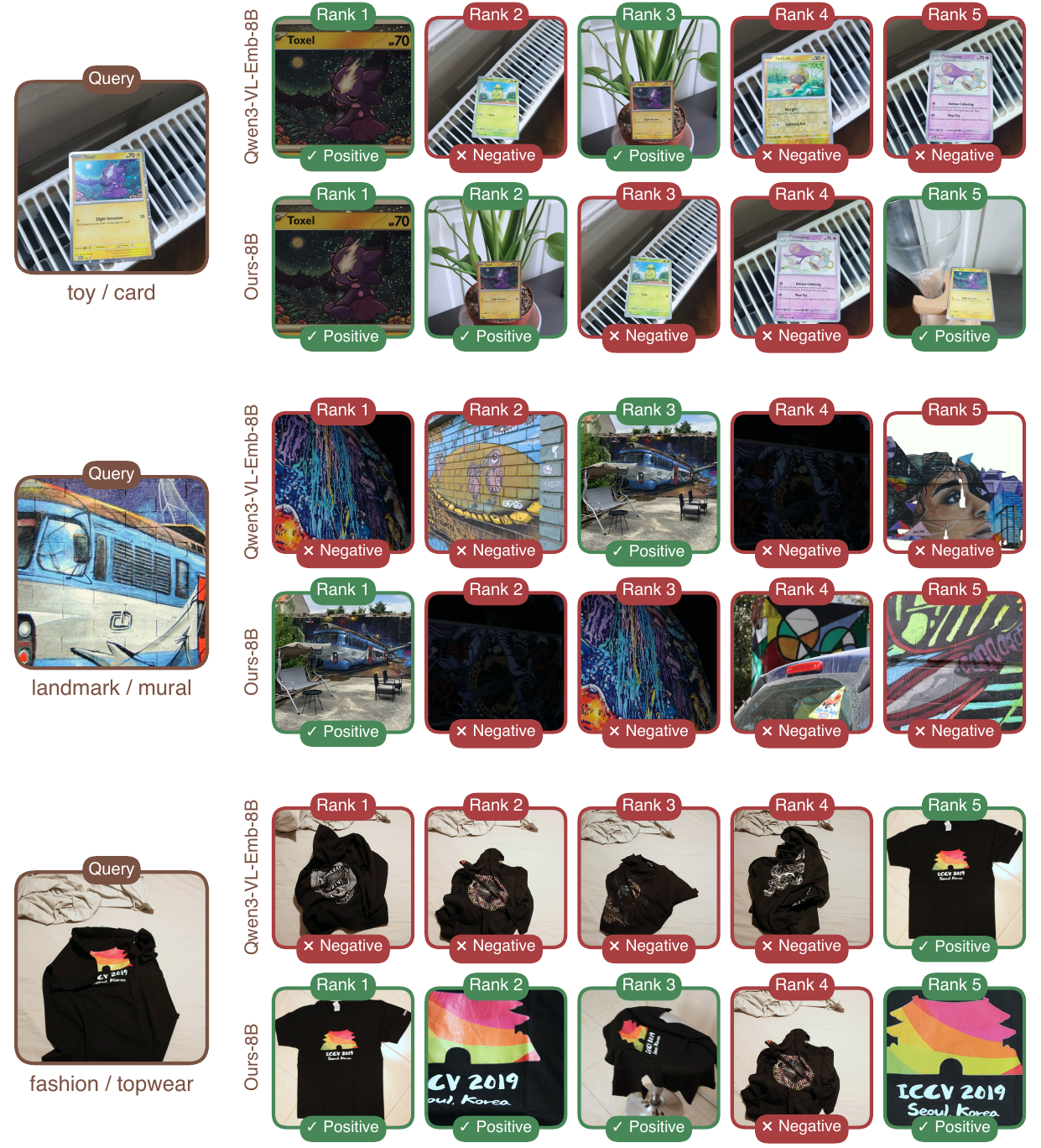}
    \caption{\textbf{Qualitative fine-grained instance retrieval on ILIAS.}
    Examples from three fine-grained classes under the category-restricted evaluation. For each query, we show the top-5 retrievals from Qwen3-VL-Embedding-8B and Ours-8B. Green positives depict the same instance as the query, while red negatives depict different instances from the same fine-grained class.}
    \label{fig:instance_retrieval_fine_grained}
\end{figure*}

Table~\ref{tab:ilias_finegrained} summarizes performance by top-level category, averaging mAP over the fine-grained classes within each category; \emph{All} averages over all $38$ classes. The finetuned model improves across every top-level category, with gains ranging from $+6.1$ to $+11.2$ points and the largest improvements on Toy ($+11.2$) and Fashion ($+10.0$).
Figure~\ref{fig:instance_retrieval_fine_grained} provides qualitative examples from this fine-grained evaluation, comparing the top-5 retrievals of the pretrained and finetuned models within individual object classes.

\begin{table}[h]
\centering
\small
\caption{
\textbf{ILIAS fine-grained instance retrieval.}
Image--image retrieval mAP (\%) averaged over the fine-grained classes within each top-level category; \emph{All} averages over all $38$ classes. Best results are shown in bold.
}
\label{tab:ilias_finegrained}

\renewcommand{\arraystretch}{1.08}
\setlength{\tabcolsep}{1.5pt}

\begin{tabular*}{\linewidth}{@{}l@{\extracolsep{\fill}}cccccccc@{}}
\toprule
\textbf{Model}
& {\footnotesize\textbf{Art}}
& {\footnotesize\textbf{Fashion}}
& {\footnotesize\textbf{Household}}
& {\footnotesize\textbf{Landmark}}
& {\footnotesize\textbf{Product}}
& {\footnotesize\textbf{Technology}}
& {\footnotesize\textbf{Toy}}
& {\footnotesize\textbf{All}} \\
\midrule

\# Classes
& $2$ & $7$ & $7$ & $4$ & $5$ & $5$ & $8$ & $38$ \\

\# Instances
& $33$ & $92$ & $50$ & $136$ & $36$ & $49$ & $95$ & $491$ \\

\midrule

Qwen3-VL-Emb-8B
& $75.4$
& $62.4$
& $72.4$
& $76.1$
& $66.0$
& $76.6$
& $52.5$
& $66.6$ \\

\textbf{Ours-8B}
& $\mathbf{82.4}$
& $\mathbf{72.4}$
& $\mathbf{80.2}$
& $\mathbf{82.2}$
& $\mathbf{72.5}$
& $\mathbf{84.4}$
& $\mathbf{63.7}$
& $\mathbf{75.2}$ \\

\bottomrule
\end{tabular*}
\end{table}

\clearpage

\clearpage

\section{Attribute Binding Beyond Attribute Swaps}
\label{app:att_binding_beyond_swaps}

SABER follows a common MCQ setup in which queries and candidates contain the same objects and attributes, with the correct candidate preserving the query's bindings and incorrect candidates reassigning attributes across objects~\citep{yuksekgonul2022and,hsieh2023sugarcrepe,bordes2023pug}. We next ask whether the resulting gains reflect broader improvements in attribute binding or specialization to this swap-based setting.

We adopt a complementary formulation from prior work~\citep{lewis2024does} in which candidates describe only part of the query. Given an image containing two attributed objects, the model retrieves a text naming a single attribute--object pair; the correct candidate preserves one of the image's bindings, whereas the hard negative candidates pair each object with the other object's attribute. Because both concepts in each candidate appear in the image, concept presence alone cannot distinguish the correct binding, and learned concept associations can even favor the mismatch~\citep{tang2023lemons}. We evaluate this formulation on the full CLEVR-based image--text benchmark introduced by \citet{lewis2024does}, combining all of its splits into a single benchmark that we denote CLEVR-bind.

\begin{figure*}[h]
    \centering
    \includegraphics[width=0.75\linewidth]{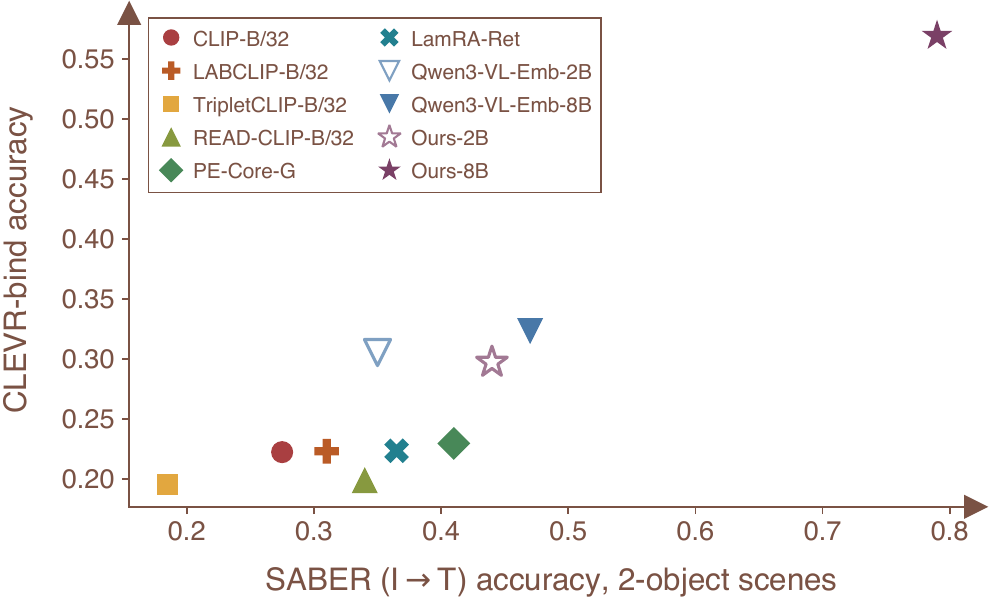}
    \caption{\textbf{Agreement across attribute-binding formulations.}
Each point represents a model, comparing SABER accuracy on two-object image--text scenes with CLEVR-Bind accuracy. Performance is positively correlated across the ten models (Spearman $\rho=0.89$; Kendall $\tau=0.78$), with Ours-8B achieving the highest accuracy in both settings.}
    \label{fig:definitions}
\end{figure*}

Since CLEVR-bind uses only two-object scenes, Figure~\ref{fig:definitions} compares model performance on CLEVR-bind with SABER in the corresponding two-object image--text setting. Across the ten evaluated models, the scores are positively correlated (Spearman $\rho=0.89$; Kendall $\tau=0.78$). Our finetuned 8B model improves over its pretrained counterpart and achieves the highest accuracy in both settings. Together, these results suggest that its gains extend beyond attribute swaps to this partial-description formulation.

\section{Can Binding-Specific Instructions Replace Finetuning?}
\label{app:instructions}
\begin{figure*}[h]
    \centering
    \includegraphics[width=0.75\linewidth]{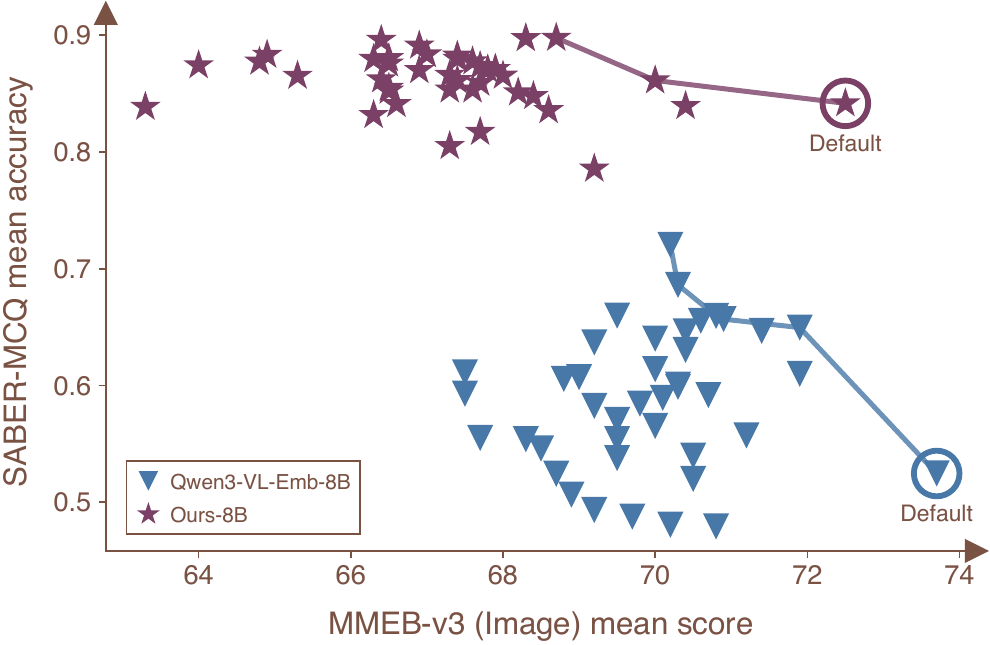}
    \caption{\textbf{Instruction trade-offs.}
Mean SABER-MCQ accuracy versus mean MMEB-v3 score on image tasks across $41$ instruction choices for Qwen3-VL-Embedding-8B and our finetuned model (Ours-8B). Lines trace each model's Pareto frontier over instructions, and circles indicate the default instruction.} \label{fig:instructions_analysis}
\end{figure*}

Qwen3-VL-Embedding conditions its embeddings on a task instruction, and its developers recommend tailoring these instructions to the target task to improve performance~\citep{qwen3vlembedding}. This raises the question of whether attribute binding can be improved through instruction choice alone, without modifying the model weights. We evaluate the default instruction (\emph{``Represent the user's input.''}) alongside $40$ binding-specific alternatives on SABER and the MMEB-v3 image tasks, using both the pretrained and finetuned 8B models. Figure~\ref{fig:instructions_analysis} shows the resulting trade-off between SABER and MMEB-v3 performance across instruction choices.

For the pretrained Qwen3-VL-Embedding-8B model, binding-specific instructions improve SABER accuracy at the expense of MMEB-v3 image performance, revealing a trade-off between the two. Although our model is finetuned only with the default instruction, it remains instruction-sensitive and exhibits a similar trade-off under the same alternatives. Crucially, with the default instruction alone, our finetuned model already surpasses the highest SABER accuracy achieved by the pretrained model under any of the $40$ binding-specific instructions. Thus, tailored instructions can further improve binding performance after finetuning, but do not replace the gains from finetuning itself.

\clearpage

\clearpage

\begin{figure*}[p]
    \centering
    \includegraphics[width=\textwidth]{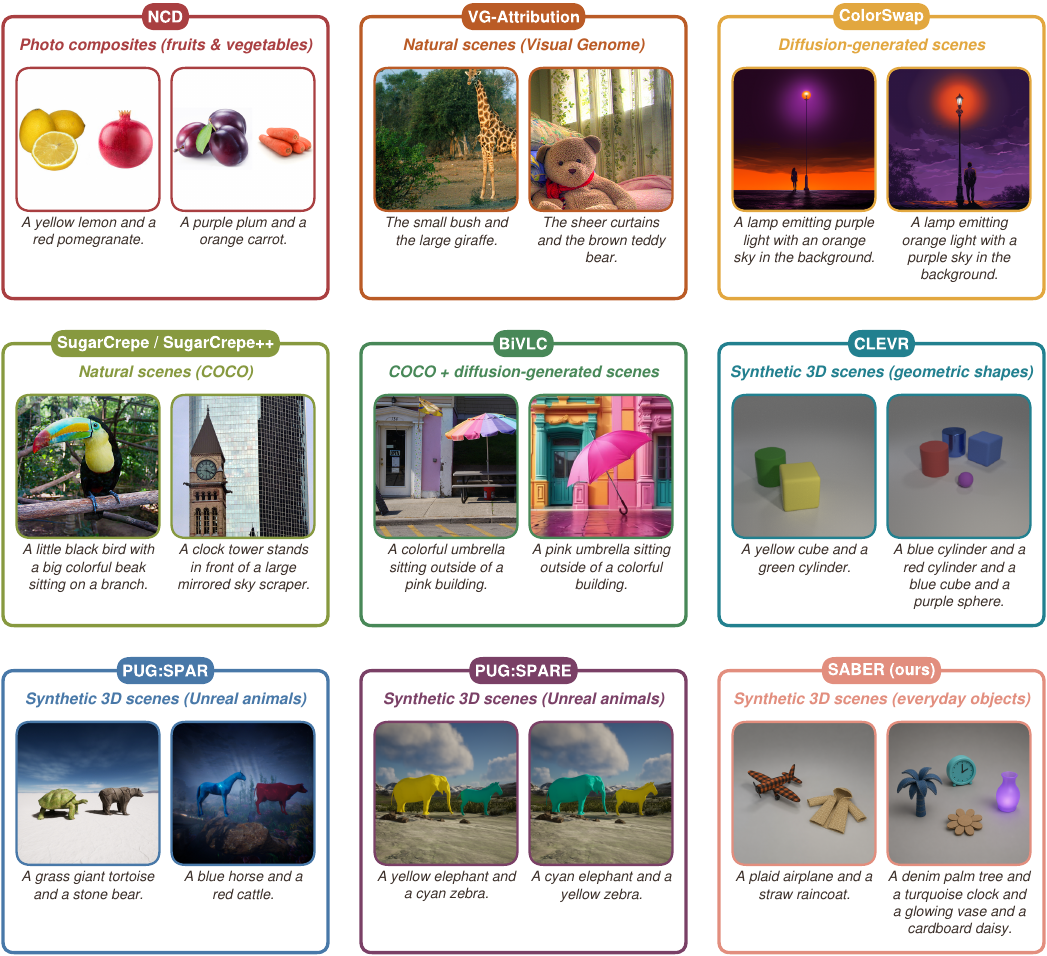}
    \caption{\textbf{Examples from attribute-binding benchmarks.}
    Existing benchmarks span natural images, photo composites, diffusion-generated scenes, and synthetic 3D environments, with varying vocabulary sizes and numbers of bindings per sample.}
    \label{fig:all_datasets}
\end{figure*}

\clearpage

\definecolor{rowgray}{HTML}{EFEEEC} 
\begin{table}[t]
\centering
\caption{
\textbf{SABER-MCQ accuracy across scene complexity.}
Accuracy (\%) for 4-way MCQs across scenes with 2--10 objects, reported for all four modality pairings. \emph{Mean} averages over object counts. Best and second-best results within each modality pairing and column are shown in bold and underlined, respectively.
}
\label{tab:saber_mcq_accuracy_full}
\resizebox{\linewidth}{!}{%
\begin{tabular}{clcccccccccc}
\toprule
 & & \multicolumn{9}{c}{\textbf{Number of objects}} & \\
\cmidrule(lr){3-11}
\textbf{Modalities} & \textbf{Model} & \textbf{2} & \textbf{3} & \textbf{4} & \textbf{5} & \textbf{6} & \textbf{7} & \textbf{8} & \textbf{9} & \textbf{10} & \textbf{Mean} \\
\midrule
\multirow{10}{*}{\textbf{T$\to$T}}
 & CLIP ViT-B/32 & 55.5 & 57.5 & 61.5 & 60.0 & 55.5 & 56.0 & 54.0 & 41.5 & 40.5 & 53.6 \\
 & \cellcolor{rowgray}LABCLIP ViT-B/32 & \cellcolor{rowgray}58.0 & \cellcolor{rowgray}57.5 & \cellcolor{rowgray}64.0 & \cellcolor{rowgray}62.5 & \cellcolor{rowgray}55.0 & \cellcolor{rowgray}57.0 & \cellcolor{rowgray}55.0 & \cellcolor{rowgray}46.5 & \cellcolor{rowgray}41.5 & \cellcolor{rowgray}55.2 \\
 & TripletCLIP ViT-B/32 & 12.0 & 36.5 & 43.5 & 48.0 & 50.5 & 49.0 & 54.0 & 43.0 & 49.5 & 42.9 \\
 & \cellcolor{rowgray}READ-CLIP ViT-B/32 & \cellcolor{rowgray}41.0 & \cellcolor{rowgray}49.5 & \cellcolor{rowgray}52.5 & \cellcolor{rowgray}55.5 & \cellcolor{rowgray}50.0 & \cellcolor{rowgray}50.5 & \cellcolor{rowgray}51.0 & \cellcolor{rowgray}39.5 & \cellcolor{rowgray}36.5 & \cellcolor{rowgray}47.3 \\
 & PE-Core-bigG-14-448 & 60.0 & 64.5 & 74.0 & 79.5 & 70.0 & 71.0 & 65.0 & 57.0 & 54.0 & 66.1 \\
 & \cellcolor{rowgray}LamRA-Ret & \cellcolor{rowgray}43.5 & \cellcolor{rowgray}53.0 & \cellcolor{rowgray}53.0 & \cellcolor{rowgray}54.0 & \cellcolor{rowgray}53.0 & \cellcolor{rowgray}53.0 & \cellcolor{rowgray}52.0 & \cellcolor{rowgray}41.5 & \cellcolor{rowgray}38.5 & \cellcolor{rowgray}49.1 \\
 & Qwen3-VL-Embedding-2B & 5.0 & 30.0 & 40.5 & 44.5 & 40.0 & 39.5 & 47.5 & 36.0 & 32.0 & 35.0 \\
 & \cellcolor{rowgray}Qwen3-VL-Embedding-8B & \cellcolor{rowgray}51.5 & \cellcolor{rowgray}56.5 & \cellcolor{rowgray}58.5 & \cellcolor{rowgray}62.5 & \cellcolor{rowgray}58.0 & \cellcolor{rowgray}53.5 & \cellcolor{rowgray}58.0 & \cellcolor{rowgray}41.5 & \cellcolor{rowgray}43.0 & \cellcolor{rowgray}53.7 \\
 & \textbf{Ours-2B} & \underline{63.0} & \underline{74.5} & \underline{81.0} & \underline{86.5} & \underline{84.5} & \underline{80.0} & \underline{80.5} & \underline{72.5} & \underline{70.0} & \underline{76.9} \\
 & \cellcolor{rowgray}\textbf{Ours-8B} & \cellcolor{rowgray}\textbf{99.0} & \cellcolor{rowgray}\textbf{98.0} & \cellcolor{rowgray}\textbf{98.5} & \cellcolor{rowgray}\textbf{99.0} & \cellcolor{rowgray}\textbf{100.0} & \cellcolor{rowgray}\textbf{99.5} & \cellcolor{rowgray}\textbf{100.0} & \cellcolor{rowgray}\textbf{100.0} & \cellcolor{rowgray}\textbf{100.0} & \cellcolor{rowgray}\textbf{99.3} \\
\midrule
\multirow{10}{*}{\textbf{T$\to$I}}
 & CLIP ViT-B/32 & 36.5 & 14.0 & 24.5 & 25.0 & 17.5 & 23.0 & 24.5 & 18.0 & 24.0 & 23.0 \\
 & \cellcolor{rowgray}LABCLIP ViT-B/32 & \cellcolor{rowgray}33.0 & \cellcolor{rowgray}17.0 & \cellcolor{rowgray}22.0 & \cellcolor{rowgray}25.5 & \cellcolor{rowgray}19.0 & \cellcolor{rowgray}20.0 & \cellcolor{rowgray}25.0 & \cellcolor{rowgray}21.0 & \cellcolor{rowgray}22.5 & \cellcolor{rowgray}22.8 \\
 & TripletCLIP ViT-B/32 & 36.0 & 30.5 & 27.5 & 28.5 & 24.0 & 23.0 & 27.5 & 22.5 & 24.0 & 27.1 \\
 & \cellcolor{rowgray}READ-CLIP ViT-B/32 & \cellcolor{rowgray}34.0 & \cellcolor{rowgray}22.0 & \cellcolor{rowgray}21.0 & \cellcolor{rowgray}21.5 & \cellcolor{rowgray}22.5 & \cellcolor{rowgray}21.5 & \cellcolor{rowgray}27.5 & \cellcolor{rowgray}22.0 & \cellcolor{rowgray}23.0 & \cellcolor{rowgray}23.9 \\
 & PE-Core-bigG-14-448 & \underline{46.0} & 29.5 & 27.5 & 27.5 & 25.5 & 26.5 & 29.0 & 26.5 & 29.5 & 29.7 \\
 & \cellcolor{rowgray}LamRA-Ret & \cellcolor{rowgray}45.5 & \cellcolor{rowgray}33.0 & \cellcolor{rowgray}27.0 & \cellcolor{rowgray}27.5 & \cellcolor{rowgray}29.0 & \cellcolor{rowgray}30.5 & \cellcolor{rowgray}25.0 & \cellcolor{rowgray}28.5 & \cellcolor{rowgray}29.5 & \cellcolor{rowgray}30.6 \\
 & Qwen3-VL-Embedding-2B & 38.0 & 28.5 & 28.5 & 22.0 & 25.5 & 29.5 & 27.0 & 27.5 & 24.0 & 27.8 \\
 & \cellcolor{rowgray}Qwen3-VL-Embedding-8B & \cellcolor{rowgray}42.5 & \cellcolor{rowgray}\underline{38.0} & \cellcolor{rowgray}33.0 & \cellcolor{rowgray}32.0 & \cellcolor{rowgray}\underline{35.5} & \cellcolor{rowgray}34.0 & \cellcolor{rowgray}33.0 & \cellcolor{rowgray}34.5 & \cellcolor{rowgray}\underline{32.0} & \cellcolor{rowgray}\underline{34.9} \\
 & \textbf{Ours-2B} & 42.0 & 34.5 & \underline{36.0} & \underline{34.5} & 26.5 & \underline{37.0} & \underline{37.5} & \underline{36.5} & 28.5 & 34.8 \\
 & \cellcolor{rowgray}\textbf{Ours-8B} & \cellcolor{rowgray}\textbf{74.0} & \cellcolor{rowgray}\textbf{66.0} & \cellcolor{rowgray}\textbf{65.0} & \cellcolor{rowgray}\textbf{68.0} & \cellcolor{rowgray}\textbf{66.5} & \cellcolor{rowgray}\textbf{71.5} & \cellcolor{rowgray}\textbf{74.0} & \cellcolor{rowgray}\textbf{69.0} & \cellcolor{rowgray}\textbf{67.0} & \cellcolor{rowgray}\textbf{69.0} \\
\midrule
\multirow{10}{*}{\textbf{I$\to$T}}
 & CLIP ViT-B/32 & 27.5 & 21.5 & 24.0 & 17.5 & 23.5 & 18.5 & 23.5 & 19.0 & 26.0 & 22.3 \\
 & \cellcolor{rowgray}LABCLIP ViT-B/32 & \cellcolor{rowgray}31.0 & \cellcolor{rowgray}27.0 & \cellcolor{rowgray}27.5 & \cellcolor{rowgray}19.5 & \cellcolor{rowgray}24.5 & \cellcolor{rowgray}23.0 & \cellcolor{rowgray}25.0 & \cellcolor{rowgray}17.5 & \cellcolor{rowgray}22.5 & \cellcolor{rowgray}24.2 \\
 & TripletCLIP ViT-B/32 & 18.5 & 24.5 & 20.0 & 27.5 & 25.0 & 17.5 & 18.0 & 23.0 & 25.5 & 22.2 \\
 & \cellcolor{rowgray}READ-CLIP ViT-B/32 & \cellcolor{rowgray}34.0 & \cellcolor{rowgray}20.5 & \cellcolor{rowgray}22.5 & \cellcolor{rowgray}16.5 & \cellcolor{rowgray}25.5 & \cellcolor{rowgray}18.5 & \cellcolor{rowgray}26.5 & \cellcolor{rowgray}23.5 & \cellcolor{rowgray}25.5 & \cellcolor{rowgray}23.7 \\
 & PE-Core-bigG-14-448 & 41.0 & 26.0 & 30.0 & 21.0 & 24.5 & 23.0 & 30.5 & 24.5 & 27.5 & 27.6 \\
 & \cellcolor{rowgray}LamRA-Ret & \cellcolor{rowgray}36.5 & \cellcolor{rowgray}25.0 & \cellcolor{rowgray}27.5 & \cellcolor{rowgray}26.0 & \cellcolor{rowgray}24.5 & \cellcolor{rowgray}28.0 & \cellcolor{rowgray}26.5 & \cellcolor{rowgray}26.0 & \cellcolor{rowgray}23.0 & \cellcolor{rowgray}27.0 \\
 & Qwen3-VL-Embedding-2B & 35.0 & 31.0 & 25.5 & 27.0 & 23.0 & 22.0 & 27.5 & 25.0 & 22.0 & 26.4 \\
 & \cellcolor{rowgray}Qwen3-VL-Embedding-8B & \cellcolor{rowgray}\underline{47.0} & \cellcolor{rowgray}36.5 & \cellcolor{rowgray}\underline{37.5} & \cellcolor{rowgray}\underline{33.0} & \cellcolor{rowgray}\underline{30.5} & \cellcolor{rowgray}\underline{33.5} & \cellcolor{rowgray}\underline{42.5} & \cellcolor{rowgray}\underline{34.0} & \cellcolor{rowgray}\underline{32.5} & \cellcolor{rowgray}\underline{36.3} \\
 & \textbf{Ours-2B} & 44.0 & \underline{38.5} & 30.5 & \underline{33.0} & 26.0 & 26.5 & 35.5 & \underline{34.0} & 28.0 & 32.9 \\
 & \cellcolor{rowgray}\textbf{Ours-8B} & \cellcolor{rowgray}\textbf{79.0} & \cellcolor{rowgray}\textbf{69.5} & \cellcolor{rowgray}\textbf{79.0} & \cellcolor{rowgray}\textbf{75.0} & \cellcolor{rowgray}\textbf{69.0} & \cellcolor{rowgray}\textbf{69.5} & \cellcolor{rowgray}\textbf{78.0} & \cellcolor{rowgray}\textbf{75.0} & \cellcolor{rowgray}\textbf{72.5} & \cellcolor{rowgray}\textbf{74.1} \\
\midrule
\multirow{10}{*}{\textbf{I$\to$I}}
 & CLIP ViT-B/32 & 73.0 & 70.0 & 72.5 & 81.5 & 71.5 & 78.0 & 79.0 & 79.0 & 75.0 & 75.5 \\
 & \cellcolor{rowgray}LABCLIP ViT-B/32 & \cellcolor{rowgray}73.0 & \cellcolor{rowgray}69.5 & \cellcolor{rowgray}73.0 & \cellcolor{rowgray}81.0 & \cellcolor{rowgray}71.5 & \cellcolor{rowgray}78.0 & \cellcolor{rowgray}78.0 & \cellcolor{rowgray}79.0 & \cellcolor{rowgray}74.5 & \cellcolor{rowgray}75.3 \\
 & TripletCLIP ViT-B/32 & 73.5 & 61.5 & 60.0 & 64.0 & 54.5 & 52.5 & 58.0 & 58.5 & 59.5 & 60.2 \\
 & \cellcolor{rowgray}READ-CLIP ViT-B/32 & \cellcolor{rowgray}83.5 & \cellcolor{rowgray}75.5 & \cellcolor{rowgray}\underline{79.5} & \cellcolor{rowgray}82.0 & \cellcolor{rowgray}81.5 & \cellcolor{rowgray}81.5 & \cellcolor{rowgray}87.5 & \cellcolor{rowgray}83.0 & \cellcolor{rowgray}86.5 & \cellcolor{rowgray}82.3 \\
 & PE-Core-bigG-14-448 & 71.0 & 67.0 & 63.0 & 67.0 & 69.5 & 68.0 & 67.0 & 69.5 & 66.0 & 67.6 \\
 & \cellcolor{rowgray}LamRA-Ret & \cellcolor{rowgray}74.5 & \cellcolor{rowgray}60.5 & \cellcolor{rowgray}61.5 & \cellcolor{rowgray}65.0 & \cellcolor{rowgray}59.5 & \cellcolor{rowgray}71.0 & \cellcolor{rowgray}67.0 & \cellcolor{rowgray}63.5 & \cellcolor{rowgray}62.0 & \cellcolor{rowgray}64.9 \\
 & Qwen3-VL-Embedding-2B & 82.5 & 67.0 & 71.0 & 77.5 & 70.0 & 76.0 & 77.5 & 71.5 & 72.5 & 73.9 \\
 & \cellcolor{rowgray}Qwen3-VL-Embedding-8B & \cellcolor{rowgray}\underline{87.0} & \cellcolor{rowgray}\underline{81.5} & \cellcolor{rowgray}79.0 & \cellcolor{rowgray}82.0 & \cellcolor{rowgray}\underline{83.0} & \cellcolor{rowgray}\underline{85.5} & \cellcolor{rowgray}\underline{89.0} & \cellcolor{rowgray}\underline{87.0} & \cellcolor{rowgray}\underline{90.0} & \cellcolor{rowgray}\underline{84.9} \\
 & \textbf{Ours-2B} & 86.0 & 77.0 & 77.0 & \underline{84.5} & 74.0 & 85.0 & 79.0 & 81.0 & 84.5 & 80.9 \\
 & \cellcolor{rowgray}\textbf{Ours-8B} & \cellcolor{rowgray}\textbf{94.5} & \cellcolor{rowgray}\textbf{94.5} & \cellcolor{rowgray}\textbf{87.5} & \cellcolor{rowgray}\textbf{91.5} & \cellcolor{rowgray}\textbf{91.0} & \cellcolor{rowgray}\textbf{97.0} & \cellcolor{rowgray}\textbf{97.0} & \cellcolor{rowgray}\textbf{98.5} & \cellcolor{rowgray}\textbf{97.0} & \cellcolor{rowgray}\textbf{94.3} \\
\bottomrule
\end{tabular}}
\end{table}

\definecolor{rowgray}{HTML}{EFEEEC} 
\begin{table}[t]
\centering
\caption{
\textbf{SABER large-gallery retrieval on independent scenes.}
mAP (\%) across scenes with 2--10 objects for the $100$ independently generated queries at each scene size, reported for all four modality pairings. \emph{Mean} averages over object counts. Best and second-best results within each modality pairing and column are shown in bold and underlined, respectively.
}
\label{tab:saber_retrieval_map_independent}
\resizebox{\linewidth}{!}{%
\begin{tabular}{clcccccccccc}
\toprule
 & & \multicolumn{9}{c}{\textbf{Number of objects}} & \\
\cmidrule(lr){3-11}
\textbf{Modalities} & \textbf{Model} & \textbf{2} & \textbf{3} & \textbf{4} & \textbf{5} & \textbf{6} & \textbf{7} & \textbf{8} & \textbf{9} & \textbf{10} & \textbf{Mean} \\
\midrule
\multirow{10}{*}{\textbf{T$\to$T}} & CLIP ViT-B/32 & 75.7 & 77.4 & 80.4 & 76.2 & 74.1 & 74.7 & 73.7 & 70.2 & 63.3 & 74.0 \\
 & \cellcolor{rowgray}LABCLIP ViT-B/32 & \cellcolor{rowgray}77.0 & \cellcolor{rowgray}76.1 & \cellcolor{rowgray}81.3 & \cellcolor{rowgray}77.0 & \cellcolor{rowgray}73.0 & \cellcolor{rowgray}72.8 & \cellcolor{rowgray}73.3 & \cellcolor{rowgray}74.2 & \cellcolor{rowgray}63.7 & \cellcolor{rowgray}74.3 \\
 & TripletCLIP ViT-B/32 & 47.7 & 59.2 & 65.5 & 67.3 & 71.8 & 67.9 & 76.5 & 67.8 & 69.7 & 65.9 \\
 & \cellcolor{rowgray}READ-CLIP ViT-B/32 & \cellcolor{rowgray}66.0 & \cellcolor{rowgray}67.1 & \cellcolor{rowgray}71.6 & \cellcolor{rowgray}75.9 & \cellcolor{rowgray}68.8 & \cellcolor{rowgray}71.1 & \cellcolor{rowgray}73.4 & \cellcolor{rowgray}64.3 & \cellcolor{rowgray}57.7 & \cellcolor{rowgray}68.4 \\
 & PE-Core-bigG-14-448 & 78.2 & 79.2 & 85.8 & 89.7 & 82.3 & 82.2 & 84.5 & 76.0 & 72.4 & 81.1 \\
 & \cellcolor{rowgray}LamRA-Ret & \cellcolor{rowgray}70.8 & \cellcolor{rowgray}71.9 & \cellcolor{rowgray}73.8 & \cellcolor{rowgray}74.7 & \cellcolor{rowgray}72.9 & \cellcolor{rowgray}71.6 & \cellcolor{rowgray}73.3 & \cellcolor{rowgray}65.0 & \cellcolor{rowgray}59.2 & \cellcolor{rowgray}70.4 \\
 & Qwen3-VL-Embedding-2B & 41.6 & 52.8 & 61.3 & 66.9 & 65.2 & 63.3 & 72.0 & 59.5 & 58.1 & 60.1 \\
 & \cellcolor{rowgray}Qwen3-VL-Embedding-8B & \cellcolor{rowgray}72.0 & \cellcolor{rowgray}74.3 & \cellcolor{rowgray}75.8 & \cellcolor{rowgray}78.6 & \cellcolor{rowgray}76.5 & \cellcolor{rowgray}74.4 & \cellcolor{rowgray}77.7 & \cellcolor{rowgray}65.7 & \cellcolor{rowgray}66.4 & \cellcolor{rowgray}73.5 \\
 & \textbf{Ours-2B} & \underline{81.3} & \underline{84.6} & \underline{89.2} & \underline{94.8} & \underline{92.4} & \underline{88.3} & \underline{90.4} & \underline{85.2} & \underline{83.3} & \underline{87.8} \\
 & \cellcolor{rowgray}\textbf{\textbf{\textbf{Ours-8B}}} & \cellcolor{rowgray}\textbf{99.0} & \cellcolor{rowgray}\textbf{98.5} & \cellcolor{rowgray}\textbf{99.5} & \cellcolor{rowgray}\textbf{99.5} & \cellcolor{rowgray}\textbf{100.0} & \cellcolor{rowgray}\textbf{99.5} & \cellcolor{rowgray}\textbf{100.0} & \cellcolor{rowgray}\textbf{100.0} & \cellcolor{rowgray}\textbf{100.0} & \cellcolor{rowgray}\textbf{99.6} \\
\midrule
\multirow{10}{*}{\textbf{T$\to$I}} & CLIP ViT-B/32 & 62.5 & 43.9 & 44.5 & 44.6 & 41.1 & 35.0 & 33.7 & 24.9 & 19.1 & 38.8 \\
 & \cellcolor{rowgray}LABCLIP ViT-B/32 & \cellcolor{rowgray}55.6 & \cellcolor{rowgray}39.1 & \cellcolor{rowgray}37.6 & \cellcolor{rowgray}44.0 & \cellcolor{rowgray}35.7 & \cellcolor{rowgray}31.9 & \cellcolor{rowgray}28.1 & \cellcolor{rowgray}23.8 & \cellcolor{rowgray}19.2 & \cellcolor{rowgray}35.0 \\
 & TripletCLIP ViT-B/32 & 26.8 & 14.2 & 11.6 & 8.3 & 5.0 & 3.3 & 3.9 & 2.1 & 3.7 & 8.8 \\
 & \cellcolor{rowgray}READ-CLIP ViT-B/32 & \cellcolor{rowgray}58.7 & \cellcolor{rowgray}45.3 & \cellcolor{rowgray}41.5 & \cellcolor{rowgray}41.3 & \cellcolor{rowgray}33.3 & \cellcolor{rowgray}29.1 & \cellcolor{rowgray}21.9 & \cellcolor{rowgray}16.5 & \cellcolor{rowgray}13.7 & \cellcolor{rowgray}33.5 \\
 & PE-Core-bigG-14-448 & 69.7 & 55.2 & 56.3 & 52.6 & 53.2 & 54.0 & 54.9 & 52.7 & 55.3 & 56.0 \\
 & \cellcolor{rowgray}LamRA-Ret & \cellcolor{rowgray}\underline{70.0} & \cellcolor{rowgray}53.8 & \cellcolor{rowgray}56.2 & \cellcolor{rowgray}58.1 & \cellcolor{rowgray}54.2 & \cellcolor{rowgray}54.9 & \cellcolor{rowgray}51.4 & \cellcolor{rowgray}54.4 & \cellcolor{rowgray}57.8 & \cellcolor{rowgray}56.8 \\
 & Qwen3-VL-Embedding-2B & 62.7 & 54.6 & 53.2 & 52.2 & 54.3 & 56.2 & 51.8 & 54.9 & 46.0 & 54.0 \\
 & \cellcolor{rowgray}Qwen3-VL-Embedding-8B & \cellcolor{rowgray}64.0 & \cellcolor{rowgray}\underline{59.6} & \cellcolor{rowgray}56.3 & \cellcolor{rowgray}\underline{60.9} & \cellcolor{rowgray}\underline{58.3} & \cellcolor{rowgray}\underline{63.5} & \cellcolor{rowgray}57.2 & \cellcolor{rowgray}57.5 & \cellcolor{rowgray}\underline{59.2} & \cellcolor{rowgray}\underline{59.6} \\
 & \textbf{Ours-2B} & 67.4 & 58.9 & \underline{58.2} & \underline{60.9} & 54.0 & 61.4 & \underline{60.1} & \underline{60.1} & 52.8 & 59.3 \\
 & \cellcolor{rowgray}\textbf{\textbf{\textbf{Ours-8B}}} & \cellcolor{rowgray}\textbf{83.0} & \cellcolor{rowgray}\textbf{81.3} & \cellcolor{rowgray}\textbf{76.8} & \cellcolor{rowgray}\textbf{82.1} & \cellcolor{rowgray}\textbf{81.9} & \cellcolor{rowgray}\textbf{83.9} & \cellcolor{rowgray}\textbf{89.0} & \cellcolor{rowgray}\textbf{81.6} & \cellcolor{rowgray}\textbf{80.7} & \cellcolor{rowgray}\textbf{82.3} \\
\midrule
\multirow{10}{*}{\textbf{I$\to$T}} & CLIP ViT-B/32 & 47.5 & 39.4 & 42.7 & 32.5 & 28.7 & 36.6 & 29.6 & 22.1 & 19.1 & 33.1 \\
 & \cellcolor{rowgray}LABCLIP ViT-B/32 & \cellcolor{rowgray}46.1 & \cellcolor{rowgray}39.8 & \cellcolor{rowgray}35.1 & \cellcolor{rowgray}29.0 & \cellcolor{rowgray}21.5 & \cellcolor{rowgray}22.8 & \cellcolor{rowgray}17.1 & \cellcolor{rowgray}9.5 & \cellcolor{rowgray}10.9 & \cellcolor{rowgray}25.8 \\
 & TripletCLIP ViT-B/32 & 14.0 & 8.4 & 5.6 & 4.0 & 3.8 & 2.2 & 2.0 & 2.1 & 3.1 & 5.0 \\
 & \cellcolor{rowgray}READ-CLIP ViT-B/32 & \cellcolor{rowgray}45.3 & \cellcolor{rowgray}31.0 & \cellcolor{rowgray}22.1 & \cellcolor{rowgray}15.3 & \cellcolor{rowgray}19.5 & \cellcolor{rowgray}23.2 & \cellcolor{rowgray}20.2 & \cellcolor{rowgray}17.3 & \cellcolor{rowgray}18.3 & \cellcolor{rowgray}23.6 \\
 & PE-Core-bigG-14-448 & 65.2 & 55.3 & 59.3 & 49.9 & 54.4 & 53.8 & 54.5 & 52.5 & 54.7 & 55.5 \\
 & \cellcolor{rowgray}LamRA-Ret & \cellcolor{rowgray}66.0 & \cellcolor{rowgray}51.7 & \cellcolor{rowgray}57.2 & \cellcolor{rowgray}51.6 & \cellcolor{rowgray}45.4 & \cellcolor{rowgray}53.7 & \cellcolor{rowgray}44.9 & \cellcolor{rowgray}51.1 & \cellcolor{rowgray}41.7 & \cellcolor{rowgray}51.5 \\
 & Qwen3-VL-Embedding-2B & 62.7 & 59.1 & 57.6 & 48.3 & 53.9 & 53.4 & 55.7 & 54.1 & 49.9 & 55.0 \\
 & \cellcolor{rowgray}Qwen3-VL-Embedding-8B & \cellcolor{rowgray}\underline{71.9} & \cellcolor{rowgray}\underline{66.6} & \cellcolor{rowgray}\underline{66.7} & \cellcolor{rowgray}\underline{59.7} & \cellcolor{rowgray}\underline{58.2} & \cellcolor{rowgray}\underline{59.0} & \cellcolor{rowgray}61.4 & \cellcolor{rowgray}59.2 & \cellcolor{rowgray}\underline{61.1} & \cellcolor{rowgray}\underline{62.6} \\
 & \textbf{Ours-2B} & 65.8 & 61.7 & 62.2 & 54.2 & 52.6 & 54.4 & \underline{62.3} & \underline{61.0} & 50.7 & 58.3 \\
 & \cellcolor{rowgray}\textbf{\textbf{\textbf{Ours-8B}}} & \cellcolor{rowgray}\textbf{84.7} & \cellcolor{rowgray}\textbf{82.8} & \cellcolor{rowgray}\textbf{87.9} & \cellcolor{rowgray}\textbf{85.3} & \cellcolor{rowgray}\textbf{79.9} & \cellcolor{rowgray}\textbf{84.2} & \cellcolor{rowgray}\textbf{89.1} & \cellcolor{rowgray}\textbf{88.8} & \cellcolor{rowgray}\textbf{87.0} & \cellcolor{rowgray}\textbf{85.5} \\
\midrule
\multirow{10}{*}{\textbf{I$\to$I}} & CLIP ViT-B/32 & 84.4 & 82.2 & 84.2 & 87.9 & 83.8 & 85.8 & 85.7 & 86.9 & 85.2 & 85.1 \\
 & \cellcolor{rowgray}LABCLIP ViT-B/32 & \cellcolor{rowgray}84.4 & \cellcolor{rowgray}82.1 & \cellcolor{rowgray}84.2 & \cellcolor{rowgray}87.9 & \cellcolor{rowgray}83.8 & \cellcolor{rowgray}85.8 & \cellcolor{rowgray}85.2 & \cellcolor{rowgray}87.1 & \cellcolor{rowgray}85.2 & \cellcolor{rowgray}85.1 \\
 & TripletCLIP ViT-B/32 & 79.9 & 74.5 & 71.7 & 68.6 & 56.1 & 54.4 & 58.5 & 55.3 & 49.7 & 63.2 \\
 & \cellcolor{rowgray}READ-CLIP ViT-B/32 & \cellcolor{rowgray}90.7 & \cellcolor{rowgray}89.5 & \cellcolor{rowgray}89.4 & \cellcolor{rowgray}88.5 & \cellcolor{rowgray}\underline{88.9} & \cellcolor{rowgray}87.7 & \cellcolor{rowgray}91.8 & \cellcolor{rowgray}89.8 & \cellcolor{rowgray}90.9 & \cellcolor{rowgray}89.7 \\
 & PE-Core-bigG-14-448 & 84.3 & 81.5 & 83.1 & 78.2 & 81.6 & 82.1 & 79.1 & 82.6 & 81.2 & 81.5 \\
 & \cellcolor{rowgray}LamRA-Ret & \cellcolor{rowgray}83.7 & \cellcolor{rowgray}76.5 & \cellcolor{rowgray}79.8 & \cellcolor{rowgray}79.1 & \cellcolor{rowgray}73.6 & \cellcolor{rowgray}84.2 & \cellcolor{rowgray}79.2 & \cellcolor{rowgray}78.7 & \cellcolor{rowgray}78.3 & \cellcolor{rowgray}79.2 \\
 & Qwen3-VL-Embedding-2B & 92.9 & 79.8 & 83.3 & 87.8 & 85.1 & 84.1 & 86.5 & 84.2 & 85.7 & 85.5 \\
 & \cellcolor{rowgray}Qwen3-VL-Embedding-8B & \cellcolor{rowgray}93.8 & \cellcolor{rowgray}\underline{90.7} & \cellcolor{rowgray}\underline{91.1} & \cellcolor{rowgray}90.7 & \cellcolor{rowgray}88.1 & \cellcolor{rowgray}\underline{92.4} & \cellcolor{rowgray}\underline{92.0} & \cellcolor{rowgray}\underline{92.7} & \cellcolor{rowgray}\underline{95.9} & \cellcolor{rowgray}\underline{91.9} \\
 & \textbf{Ours-2B} & \underline{94.9} & 87.2 & 87.5 & \underline{92.4} & 88.5 & 89.3 & 86.2 & 90.1 & 92.4 & 89.8 \\
 & \cellcolor{rowgray}\textbf{\textbf{\textbf{Ours-8B}}} & \cellcolor{rowgray}\textbf{96.3} & \cellcolor{rowgray}\textbf{97.2} & \cellcolor{rowgray}\textbf{94.9} & \cellcolor{rowgray}\textbf{94.2} & \cellcolor{rowgray}\textbf{94.4} & \cellcolor{rowgray}\textbf{98.5} & \cellcolor{rowgray}\textbf{98.5} & \cellcolor{rowgray}\textbf{100.0} & \cellcolor{rowgray}\textbf{99.5} & \cellcolor{rowgray}\textbf{97.1} \\
\bottomrule
\end{tabular}}
\end{table}

\begin{table}[t]
\centering
\caption{
\textbf{SABER large-gallery retrieval on nested chains.}
mAP (\%) across scenes with 2--10 objects for the $100$ nested-chain queries at each scene size, reported for all four modality pairings. \emph{Mean} averages over object counts. Best and second-best results within each modality pairing and column are shown in bold and underlined, respectively.
}
\label{tab:saber_retrieval_map_nested}
\resizebox{\linewidth}{!}{%
\begin{tabular}{clcccccccccc}
\toprule
 & & \multicolumn{9}{c}{\textbf{Number of objects}} & \\
\cmidrule(lr){3-11}
\textbf{Modalities} & \textbf{Model} & \textbf{2} & \textbf{3} & \textbf{4} & \textbf{5} & \textbf{6} & \textbf{7} & \textbf{8} & \textbf{9} & \textbf{10} & \textbf{Mean} \\
\midrule
\multirow{10}{*}{\textbf{T$\to$T}} & CLIP ViT-B/32 & 25.0 & 28.6 & 30.7 & 33.1 & 38.8 & 39.6 & 42.2 & 46.6 & 57.4 & 38.0 \\
 & \cellcolor{rowgray}LABCLIP ViT-B/32 & \cellcolor{rowgray}27.2 & \cellcolor{rowgray}30.0 & \cellcolor{rowgray}32.1 & \cellcolor{rowgray}34.1 & \cellcolor{rowgray}37.6 & \cellcolor{rowgray}39.0 & \cellcolor{rowgray}42.4 & \cellcolor{rowgray}43.7 & \cellcolor{rowgray}57.0 & \cellcolor{rowgray}38.1 \\
 & TripletCLIP ViT-B/32 & 18.8 & 25.5 & 28.3 & 31.1 & 33.3 & 36.2 & 41.8 & 45.3 & 62.1 & 35.8 \\
 & \cellcolor{rowgray}READ-CLIP ViT-B/32 & \cellcolor{rowgray}26.4 & \cellcolor{rowgray}30.0 & \cellcolor{rowgray}31.2 & \cellcolor{rowgray}33.1 & \cellcolor{rowgray}36.9 & \cellcolor{rowgray}38.7 & \cellcolor{rowgray}41.0 & \cellcolor{rowgray}44.1 & \cellcolor{rowgray}57.8 & \cellcolor{rowgray}37.7 \\
 & PE-Core-bigG-14-448 & 30.1 & 32.9 & 34.5 & 36.3 & 38.6 & 43.0 & 44.6 & 50.2 & 63.7 & 41.5 \\
 & \cellcolor{rowgray}LamRA-Ret & \cellcolor{rowgray}25.0 & \cellcolor{rowgray}30.3 & \cellcolor{rowgray}31.5 & \cellcolor{rowgray}34.5 & \cellcolor{rowgray}37.0 & \cellcolor{rowgray}39.3 & \cellcolor{rowgray}39.4 & \cellcolor{rowgray}43.0 & \cellcolor{rowgray}56.0 & \cellcolor{rowgray}37.3 \\
 & Qwen3-VL-Embedding-2B & 24.7 & 29.0 & 31.0 & 32.5 & 33.6 & 34.2 & 37.0 & 38.6 & 45.5 & 34.0 \\
 & \cellcolor{rowgray}Qwen3-VL-Embedding-8B & \cellcolor{rowgray}29.1 & \cellcolor{rowgray}31.8 & \cellcolor{rowgray}33.5 & \cellcolor{rowgray}36.5 & \cellcolor{rowgray}39.4 & \cellcolor{rowgray}39.4 & \cellcolor{rowgray}41.6 & \cellcolor{rowgray}39.8 & \cellcolor{rowgray}49.3 & \cellcolor{rowgray}37.8 \\
 & \textbf{Ours-2B} & \underline{31.8} & \underline{35.4} & \underline{38.7} & \underline{42.2} & \underline{45.0} & \underline{47.4} & \underline{50.5} & \underline{58.3} & \underline{66.4} & \underline{46.2} \\
 & \cellcolor{rowgray}\textbf{\textbf{\textbf{Ours-8B}}} & \cellcolor{rowgray}\textbf{44.5} & \cellcolor{rowgray}\textbf{48.7} & \cellcolor{rowgray}\textbf{53.5} & \cellcolor{rowgray}\textbf{59.6} & \cellcolor{rowgray}\textbf{65.7} & \cellcolor{rowgray}\textbf{72.4} & \cellcolor{rowgray}\textbf{77.3} & \cellcolor{rowgray}\textbf{81.0} & \cellcolor{rowgray}\textbf{86.4} & \cellcolor{rowgray}\textbf{65.5} \\
\midrule
\multirow{10}{*}{\textbf{T$\to$I}} & CLIP ViT-B/32 & 13.6 & 15.0 & 15.3 & 14.5 & 13.8 & 13.6 & 12.5 & 11.4 & 13.2 & 13.7 \\
 & \cellcolor{rowgray}LABCLIP ViT-B/32 & \cellcolor{rowgray}12.1 & \cellcolor{rowgray}14.1 & \cellcolor{rowgray}14.0 & \cellcolor{rowgray}13.6 & \cellcolor{rowgray}12.9 & \cellcolor{rowgray}11.5 & \cellcolor{rowgray}13.4 & \cellcolor{rowgray}11.7 & \cellcolor{rowgray}13.9 & \cellcolor{rowgray}13.0 \\
 & TripletCLIP ViT-B/32 & 5.4 & 6.2 & 6.0 & 5.7 & 5.9 & 5.6 & 5.4 & 3.5 & 5.1 & 5.4 \\
 & \cellcolor{rowgray}READ-CLIP ViT-B/32 & \cellcolor{rowgray}11.5 & \cellcolor{rowgray}12.8 & \cellcolor{rowgray}12.2 & \cellcolor{rowgray}12.6 & \cellcolor{rowgray}11.6 & \cellcolor{rowgray}9.5 & \cellcolor{rowgray}10.7 & \cellcolor{rowgray}7.7 & \cellcolor{rowgray}7.6 & \cellcolor{rowgray}10.7 \\
 & PE-Core-bigG-14-448 & 26.7 & 26.1 & 24.4 & 22.8 & 21.8 & 21.1 & 22.9 & 21.7 & 24.9 & 23.6 \\
 & \cellcolor{rowgray}LamRA-Ret & \cellcolor{rowgray}19.5 & \cellcolor{rowgray}24.0 & \cellcolor{rowgray}23.3 & \cellcolor{rowgray}24.3 & \cellcolor{rowgray}26.0 & \cellcolor{rowgray}26.7 & \cellcolor{rowgray}28.6 & \cellcolor{rowgray}34.5 & \cellcolor{rowgray}39.3 & \cellcolor{rowgray}27.3 \\
 & Qwen3-VL-Embedding-2B & 23.3 & 26.4 & 27.1 & 26.6 & 25.4 & 26.9 & 30.6 & 31.2 & 35.4 & 28.1 \\
 & \cellcolor{rowgray}Qwen3-VL-Embedding-8B & \cellcolor{rowgray}\underline{28.3} & \cellcolor{rowgray}\underline{30.6} & \cellcolor{rowgray}\underline{30.1} & \cellcolor{rowgray}\underline{30.8} & \cellcolor{rowgray}\underline{32.8} & \cellcolor{rowgray}\underline{30.9} & \cellcolor{rowgray}\underline{37.0} & \cellcolor{rowgray}\underline{37.4} & \cellcolor{rowgray}\underline{40.2} & \cellcolor{rowgray}\underline{33.1} \\
 & \textbf{Ours-2B} & 26.3 & 28.4 & 28.9 & 28.3 & 25.9 & 26.6 & 30.0 & 34.8 & 35.5 & 29.4 \\
 & \cellcolor{rowgray}\textbf{\textbf{\textbf{Ours-8B}}} & \cellcolor{rowgray}\textbf{35.1} & \cellcolor{rowgray}\textbf{36.3} & \cellcolor{rowgray}\textbf{37.7} & \cellcolor{rowgray}\textbf{39.6} & \cellcolor{rowgray}\textbf{43.6} & \cellcolor{rowgray}\textbf{46.0} & \cellcolor{rowgray}\textbf{54.3} & \cellcolor{rowgray}\textbf{61.7} & \cellcolor{rowgray}\textbf{62.9} & \cellcolor{rowgray}\textbf{46.3} \\
\midrule
\multirow{10}{*}{\textbf{I$\to$T}} & CLIP ViT-B/32 & 15.7 & 16.3 & 17.0 & 17.2 & 17.5 & 16.3 & 16.3 & 14.5 & 14.3 & 16.1 \\
 & \cellcolor{rowgray}LABCLIP ViT-B/32 & \cellcolor{rowgray}13.0 & \cellcolor{rowgray}12.8 & \cellcolor{rowgray}13.9 & \cellcolor{rowgray}13.3 & \cellcolor{rowgray}11.9 & \cellcolor{rowgray}11.4 & \cellcolor{rowgray}12.5 & \cellcolor{rowgray}11.2 & \cellcolor{rowgray}9.1 & \cellcolor{rowgray}12.1 \\
 & TripletCLIP ViT-B/32 & 4.4 & 4.8 & 3.6 & 3.7 & 3.5 & 3.3 & 3.9 & 2.2 & 2.3 & 3.5 \\
 & \cellcolor{rowgray}READ-CLIP ViT-B/32 & \cellcolor{rowgray}10.3 & \cellcolor{rowgray}12.0 & \cellcolor{rowgray}12.7 & \cellcolor{rowgray}13.8 & \cellcolor{rowgray}14.8 & \cellcolor{rowgray}13.2 & \cellcolor{rowgray}13.6 & \cellcolor{rowgray}14.4 & \cellcolor{rowgray}13.5 & \cellcolor{rowgray}13.1 \\
 & PE-Core-bigG-14-448 & \underline{29.7} & \underline{28.9} & \underline{29.5} & 29.3 & \underline{30.9} & \underline{32.1} & \underline{37.9} & 36.7 & 40.9 & \underline{32.9} \\
 & \cellcolor{rowgray}LamRA-Ret & \cellcolor{rowgray}22.1 & \cellcolor{rowgray}25.0 & \cellcolor{rowgray}25.5 & \cellcolor{rowgray}26.7 & \cellcolor{rowgray}28.0 & \cellcolor{rowgray}28.7 & \cellcolor{rowgray}31.2 & \cellcolor{rowgray}30.1 & \cellcolor{rowgray}34.6 & \cellcolor{rowgray}28.0 \\
 & Qwen3-VL-Embedding-2B & 24.6 & 27.2 & 27.6 & \underline{30.6} & 30.2 & 29.9 & 34.2 & 35.4 & 38.1 & 30.9 \\
 & \cellcolor{rowgray}Qwen3-VL-Embedding-8B & \cellcolor{rowgray}26.2 & \cellcolor{rowgray}27.5 & \cellcolor{rowgray}28.2 & \cellcolor{rowgray}28.3 & \cellcolor{rowgray}28.9 & \cellcolor{rowgray}27.9 & \cellcolor{rowgray}34.3 & \cellcolor{rowgray}31.2 & \cellcolor{rowgray}36.6 & \cellcolor{rowgray}29.9 \\
 & \textbf{Ours-2B} & 27.5 & 28.6 & 28.6 & 30.0 & 30.1 & 31.3 & 33.1 & \underline{40.7} & \underline{42.8} & 32.5 \\
 & \cellcolor{rowgray}\textbf{\textbf{\textbf{Ours-8B}}} & \cellcolor{rowgray}\textbf{34.1} & \cellcolor{rowgray}\textbf{34.9} & \cellcolor{rowgray}\textbf{37.6} & \cellcolor{rowgray}\textbf{39.3} & \cellcolor{rowgray}\textbf{43.2} & \cellcolor{rowgray}\textbf{44.8} & \cellcolor{rowgray}\textbf{52.3} & \cellcolor{rowgray}\textbf{59.1} & \cellcolor{rowgray}\textbf{68.2} & \cellcolor{rowgray}\textbf{46.0} \\
\midrule
\multirow{10}{*}{\textbf{I$\to$I}} & CLIP ViT-B/32 & 21.6 & 29.3 & 37.8 & \underline{47.7} & 51.6 & 58.4 & 65.7 & 66.4 & 65.0 & 49.3 \\
 & \cellcolor{rowgray}LABCLIP ViT-B/32 & \cellcolor{rowgray}21.6 & \cellcolor{rowgray}29.2 & \cellcolor{rowgray}37.8 & \cellcolor{rowgray}\underline{47.7} & \cellcolor{rowgray}51.7 & \cellcolor{rowgray}58.4 & \cellcolor{rowgray}65.5 & \cellcolor{rowgray}66.4 & \cellcolor{rowgray}64.5 & \cellcolor{rowgray}49.2 \\
 & TripletCLIP ViT-B/32 & 17.1 & 20.6 & 24.8 & 31.4 & 30.8 & 36.5 & 39.2 & 46.3 & 42.0 & 32.1 \\
 & \cellcolor{rowgray}READ-CLIP ViT-B/32 & \cellcolor{rowgray}24.3 & \cellcolor{rowgray}31.6 & \cellcolor{rowgray}\underline{41.1} & \cellcolor{rowgray}\textbf{50.8} & \cellcolor{rowgray}\textbf{56.4} & \cellcolor{rowgray}\underline{62.2} & \cellcolor{rowgray}\underline{69.9} & \cellcolor{rowgray}73.6 & \cellcolor{rowgray}76.6 & \cellcolor{rowgray}\underline{54.1} \\
 & PE-Core-bigG-14-448 & 30.0 & 33.2 & 34.1 & 39.5 & 43.8 & 49.3 & 58.1 & 65.3 & 70.5 & 47.1 \\
 & \cellcolor{rowgray}LamRA-Ret & \cellcolor{rowgray}26.0 & \cellcolor{rowgray}30.6 & \cellcolor{rowgray}32.9 & \cellcolor{rowgray}36.4 & \cellcolor{rowgray}39.5 & \cellcolor{rowgray}44.3 & \cellcolor{rowgray}51.6 & \cellcolor{rowgray}60.6 & \cellcolor{rowgray}71.2 & \cellcolor{rowgray}43.7 \\
 & Qwen3-VL-Embedding-2B & 28.5 & 32.7 & 35.6 & 37.8 & 42.6 & 48.8 & 56.7 & 66.8 & 76.2 & 47.3 \\
 & \cellcolor{rowgray}Qwen3-VL-Embedding-8B & \cellcolor{rowgray}\underline{32.4} & \cellcolor{rowgray}\underline{35.5} & \cellcolor{rowgray}36.5 & \cellcolor{rowgray}40.2 & \cellcolor{rowgray}46.4 & \cellcolor{rowgray}52.0 & \cellcolor{rowgray}62.9 & \cellcolor{rowgray}\underline{76.3} & \cellcolor{rowgray}\underline{88.7} & \cellcolor{rowgray}52.3 \\
 & \textbf{Ours-2B} & 31.4 & 34.2 & 36.7 & 39.3 & 42.1 & 51.2 & 59.6 & 70.5 & 83.9 & 49.9 \\
 & \cellcolor{rowgray}\textbf{\textbf{\textbf{Ours-8B}}} & \cellcolor{rowgray}\textbf{37.2} & \cellcolor{rowgray}\textbf{39.6} & \cellcolor{rowgray}\textbf{41.4} & \cellcolor{rowgray}45.9 & \cellcolor{rowgray}\underline{54.0} & \cellcolor{rowgray}\textbf{63.1} & \cellcolor{rowgray}\textbf{72.1} & \cellcolor{rowgray}\textbf{83.5} & \cellcolor{rowgray}\textbf{90.4} & \cellcolor{rowgray}\textbf{58.6} \\
\bottomrule
\end{tabular}}
\end{table}

\begin{table}[h]
\centering
\small
\renewcommand{\arraystretch}{1.16}
\caption{\textbf{Licenses and usage terms of datasets and models used in this work.}}
\label{tab:licenses}
\begin{tabular*}{\linewidth}{@{\extracolsep{\fill}}llll@{}}
\toprule
\textbf{Dataset}
& \textbf{License}
& \textbf{Model}
& \textbf{License} \\
\midrule

NCD
& MIT
& CLIP ViT-B/32
& MIT \\
\addlinespace[1.5pt]

VG-Attribution
& MIT
& LABCLIP ViT-B/32
& MIT \\
\addlinespace[1.5pt]

ColorSwap
& MIT
& TripletCLIP ViT-B/32
& Open \\
\addlinespace[1.5pt]

SugarCrepe-Swap-Att.
& MIT
& READ-CLIP ViT-B/32
& MIT \\
\addlinespace[1.5pt]

SugarCrepe++-Swap-Att.
& MIT
& PE-Core bigG-14-448
& Apache 2.0 \\
\addlinespace[1.5pt]

BiVLC-Swap-Att.
& MIT
& LamRA-Ret
& MIT \\
\addlinespace[1.5pt]

CLEVR~\citep{koishigarina2025clip}
& CC BY 4.0
& Qwen3-VL-Embedding
& Apache 2.0 \\
\addlinespace[1.5pt]

PUG:SPAR
& CC BY-NC 4.0
& Qwen3-VL-Reranker
& Apache 2.0 \\
\addlinespace[1.5pt]

PUG:SPARE
& CC BY-NC 4.0
& Gemma 4 31B-IT
& Apache 2.0 \\
\addlinespace[1.5pt]

MMEB-v3
& MIT
& GPT-Image-2
& Proprietary \\
\addlinespace[1.5pt]

CLEVR~\citep{lewis2024does}
& BSD-3-Clause
& WeMM-Embedding
& Apache-2.0 \\
\addlinespace[1.5pt]

NegBench
& MIT
& UEmbed
& CC BY 4.0 \\
\addlinespace[1.5pt]

ILIAS
& MIT
& 
& \\
\addlinespace[1.5pt]

INSTRE
& Flickr T\&C
&
& \\
\addlinespace[1.5pt]

PODS
& MIT
&
& \\
\addlinespace[1.5pt]

Paris6K
& Flickr T\&C + NC 
&
& \\
\addlinespace[1.5pt]

Oxford5K
& Flickr T\&C + NC 
&
& \\
\addlinespace[1.5pt]

Sculptures6K
& Flickr T\&C + NC 
&
& \\

\bottomrule
\end{tabular*}
\end{table}

\end{document}